\documentclass{bmvc2k}

\title{Improving Classification of Occluded
Objects through Scene Context}

\addauthor{Courtney M. King}{cking74@fordham.edu}{1}
\addauthor{Daniel D. Leeds}{dleeds@fordham.edu}{1}
\addauthor{Damian Lyons}{dlyons@fordham.edu}{1}
\addauthor{George Kalaitzis}{gkalaitzis@fordham.edu}{1}

\addinstitution{
Fordham University \\ 
Department of Computer Science \\ 
Bronx, NY, 10458 
}

\runninghead{}{}

\begin{document}

\maketitle

\begin{abstract}
The presence of occlusions has provided substantial challenges to typically-powerful object recognition algorithms.  
Additional sources of information can be extremely valuable to reduce errors caused by occlusions. 
Scene context is known to aid in object recognition in biological vision.
In this work, we attempt to add robustness into existing Region Proposal Network-Deep Convolutional Neural Network (RPN-DCNN) object detection networks through two distinct scene-based information fusion techniques. 
We present one algorithm under each methodology:
the first operates prior to prediction,
selecting a custom object network to use based on the identified background scene,
and the second operates after detection,
fusing scene knowledge into initial object scores output by the RPN.
We demonstrate our algorithms on challenging datasets featuring partial occlusions, which show overall improvement in both recall and precision against baseline methods. 
In addition, our experiments contrast multiple training methodologies for occlusion handling, finding that training on a combination of both occluded and unoccluded images demonstrates an improvement over the others. 
Our method is interpretable and can easily be adapted to other datasets, offering many future directions for research and practical applications.
\end{abstract}

\section{Introduction}\label{sec:intro}

\begin{figure}
    \includegraphics[width=0.24 \linewidth]{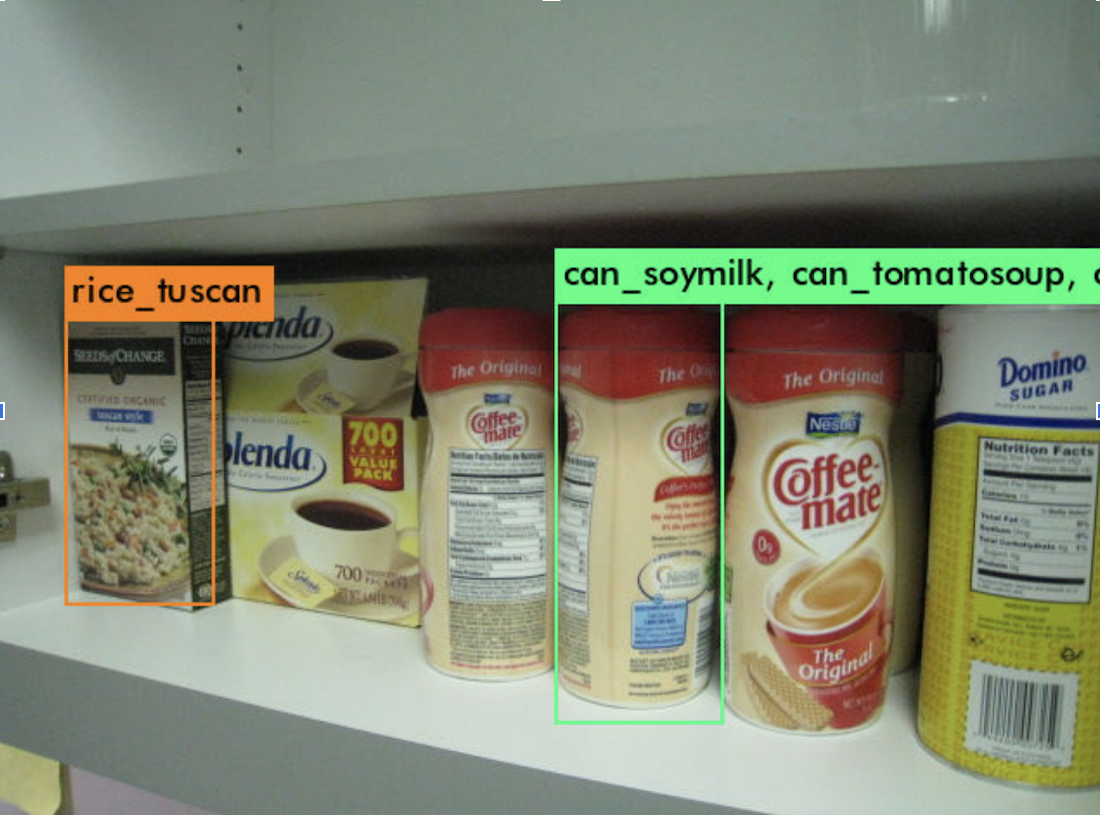}
    \includegraphics[width=0.24 \linewidth]{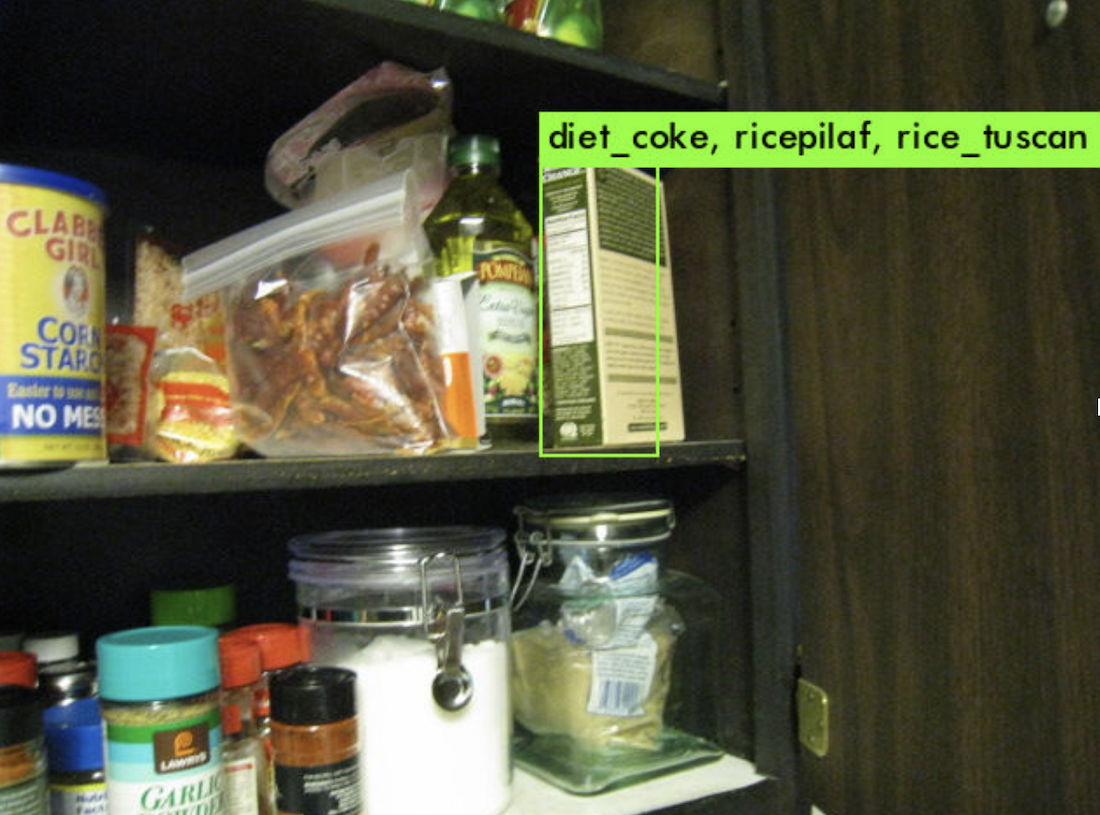}
    \includegraphics[width=0.24 \linewidth]{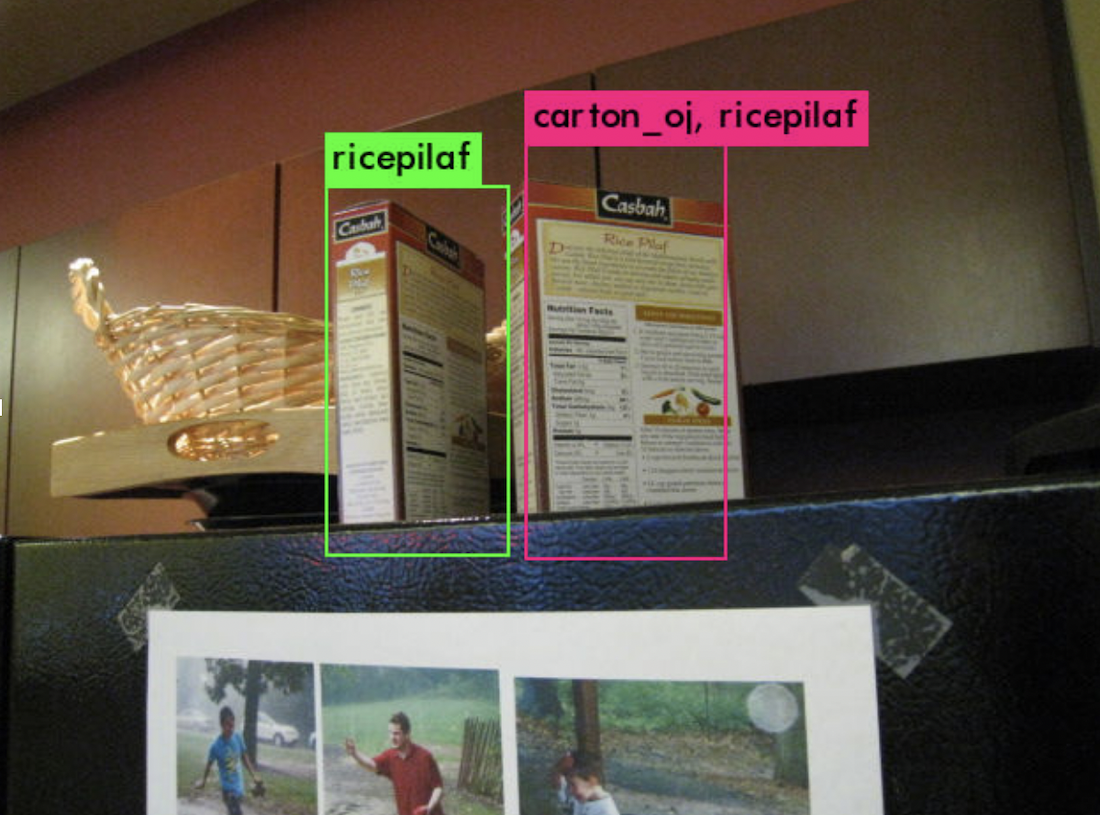}
    \includegraphics[width=0.24 \linewidth]{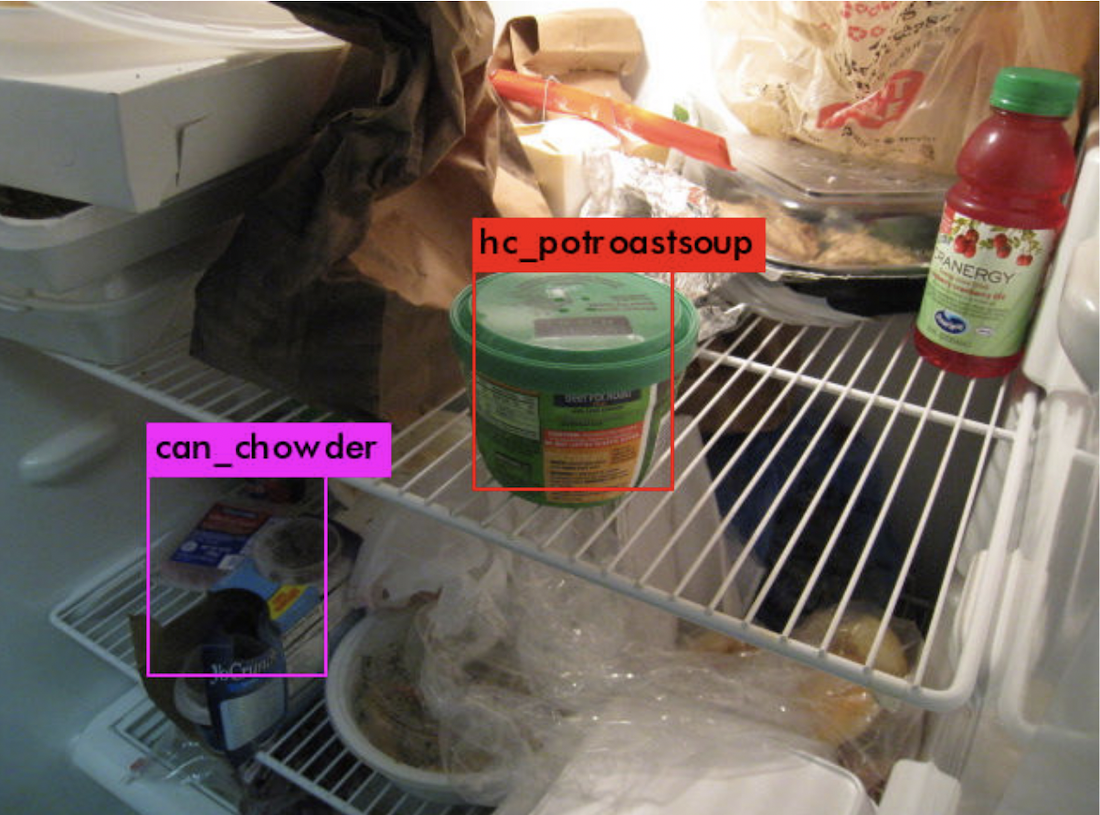} 
    \\
    \caption{Misclassified objects in Occluded Groceries dataset that can be resolved using scene context.}
    \label{fig:yolo_fails}
\end{figure}

Object detection and classification is a popular topic in computer vision with many useful applications ranging from target tracking to autopilot software \cite{lin2014microsoft} \cite{hwang2015multispectral}.  Modern computer vision is well-equipped to handle these applications in general, however, it still faces challenges, such as cases when target objects are not directly in view or upon the presence of occlusions.  Occlusions are full or partial blockages, which can impair the detection or classification of objects.  Certain occlusion types are prone to arise in object detection, including shadows or blockage by other objects \cite{ruan2023review} \cite{zhang2019pose2seg}. 
Handling these occlusions has remained an active area of research given the necessity for robust object recognition models \cite{zhan2210tri} \cite{kassaw2024deep}.

One way to address the problem is by incorporating additional sources of information. 
Prior distributions are useful to object detection in a multitude of ways ranging from generative models to statistical methods.
Generative models can be used to learn the shape of objects, leading to corrective procedures such as reconstruction \cite{6247702}.
On the other hand, statistical information can be used to give greater reasoning capacity based on context.
Context, broadly defined as additional information about or within the image, has proven useful to a variety of computer and human vision tasks \cite{wang2023context}.
It can be leveraged by extracting features from the greater image to combine with prior knowledge. 
Context exists at several aspects of an image which aid in detection tasks \cite{wang2024gmc}.
It exists between objects \cite{aminoff2022contextual} and geometric knowledge and relationships can be used correct object detection mistakes \cite{hoiem2008putting}.
It also exists at the scene-level \cite{xiao2016sun}, which can similarly be used for adjacent tasks, such as detecting 
out-of-context objects \cite{bomatter2021pigs} or inferring missing objects from scenes \cite{sun2017seeing}.

This project explores the use of scene context to improve object classification in the presence of occluders. Specifically, we use scene context from the images to obtain scene classes. 
The first algorithm implements multiple scene-centric networks and the second incorporates the object-scene co-occurrence statistics. We test our methods on challenging datasets featuring occluded grocery items \cite{hsiao2010making} and litter in the wild \cite{taco2020}. Our results demonstrate a reasonable performance gain, showing increased scores in both average recall and precision, as well as within a majority of the individual object categories.

\section{Related Works}\label{sec:background}
Object detection is a two-step procedure including both a localization and classification step. The detection step involves locating individual objects in the larger frame. A common way this is performed is by Region Proposal Networks (RPNs). 
These work by predicting a series of varying bounding boxes within an image, then each box is given a score representing the likelihood of that region containing an object. The boxes receiving a score below a specified threshold are discarded. For those not discarded, the spatial information saved.
Some popular models for detection tasks include Fast R-CNN\cite{girshick2015fast} Faster R-CNN\cite{ren2016faster} and Mask R-CNN\cite{he2017mask}. 
Secondly, the classification step assigns a class label to the regions extracted from the image in the initial step. There are also various methods that can be used in this stage, with many based on Deep Convolutional Neural Network (DCNN) implementations, including common models like VGG-16\cite{simonyan2014very} and ResNet\cite{he2016deep}. 
In addition to these two models classes, there are also "one-shot" detectors which handle both steps simultaneously. In this category exists the various YOLO models\cite{diwan2023object}.  YOLO excels at real-time object detection, which is useful for tasks in robotics and manufacturing. It does this using a single CNN network over the entire image. An input image is divided into an $NxN$ grid, and the grid square on which the center of objects lay are responsible for the detection of the object. Thus, each grid square predicts bounding boxes together with the class probabilities and confidence scores in a single pass. This gives the model an edge of speed, however, its localization performance may be less precise, and it may fail to predict certain objects accurately, such as those which are small or feature occlusions.

Occlusions are a persistent problem in object recognition, spurring the development of more robust models. Common object detection models which use a DCNN for classification often exhibit a decrease in performance based on the level of occlusion \cite{kortylewski2021compositional} \cite{CompNet:CVPR:2020}.
RPNs struggle to predict accurate bounding boxes in severe occlusions, which subsequently impact the performance of the DCNNs. 
Methods of occlusion handling can exist at various stages in the process. 
As the representation of the objects can affect the bounding-box prediction, parts-based models can be useful for tighter or more accurate detections\cite{fischler1973representation}. 
The key point of these models has to do with incorporating both local and global information into a classification score, or similarly of the use both semantic and syntactic information in the image. While these models are effective, their computational cost makes them less than ideal for real-time object detection. However, solutions for improving light-weight object detection networks have been proposed. Integrating temporal context into Faster RCNN for occluded object detection has shown an improvement in video streams \cite{yap2024improving}, while YOLO-ACN was designed to improve the detection of small or occluded objects \cite{li2020yolo}.

Modern computer vision offers a breadth of context-based models and solutions \cite{wang2023context}. 
It has been shown that contextual information appears encoded when performing recognition on objects on a white background when using a network trained on recognizing objects in their typical contexts\cite{aminoff2022contextual}. 
In some models, context may exist and be used implicitly, however, others have been developed which explicitly incorporate context into their design. The use of context may be limited to a single task or incorporated throughout by multi-stage context models \cite{wang2024gmc} \cite{wang2022multiclu}. In recent years, models have been developed to solve more specific problems in object detection, such as Context RNN \cite{beery2020context}, which uses temporal context to improve object detection during long observational windows. 
Context can be incorporated into classic models in various ways to achieve occlusion handling. 
One recent example is CompositionalNets, which work by localizing occluders and incorporating non-occluded regions of the image into the classification. 
While these have shown an improvement over DCNNs in classifying partially occluded objects, they still struggle in heavy occlusions and are less discriminative at classifying non-occluded objects\cite{CompNet:CVPR:2020}. 
Another similar example is Context-aware CompositionalNets (CACN) \cite{kortylewski2021compositional} which can be understood as a two-step process in which an additional layer is added before the classification step that first localizes and masks the occluders.  It combines both DCNNs and CompositionalNets to offer a more generalizable classification model in the case of partially occluded objects. 

\section{Methodology}\label{sec:methodology}
% This section summarizes several context-based occlusion-handling techniques for improved object detection and classification on challenging datasets. 

% Towards this aim, we discuss two distinct approaches, which both incorporate scene context, albeit at different stages, and offer two algorithms for implementation. The first approach targets the pre-detection stage of custom-trained networks. The second approach targets the post-detection stage. We also include three distinct training methodologies, which exhibit a significant difference in performance from each other, for testing our models and baselines on occluded images.

\subsection{Scene Context}

\begin{figure}[h!]
    \includegraphics[width=0.24 \linewidth]{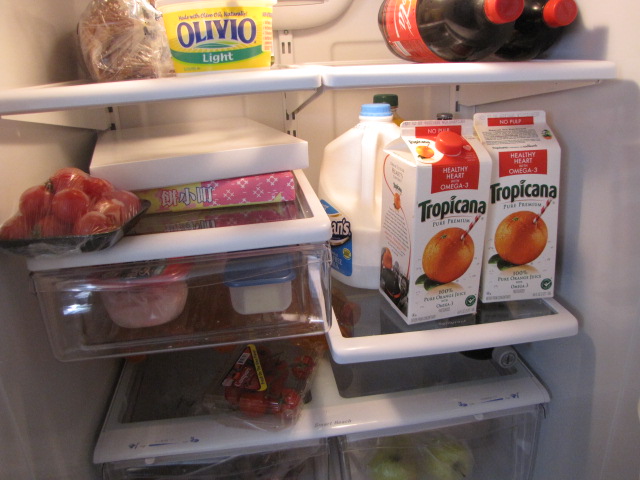}
    \includegraphics[width=0.24 \linewidth]{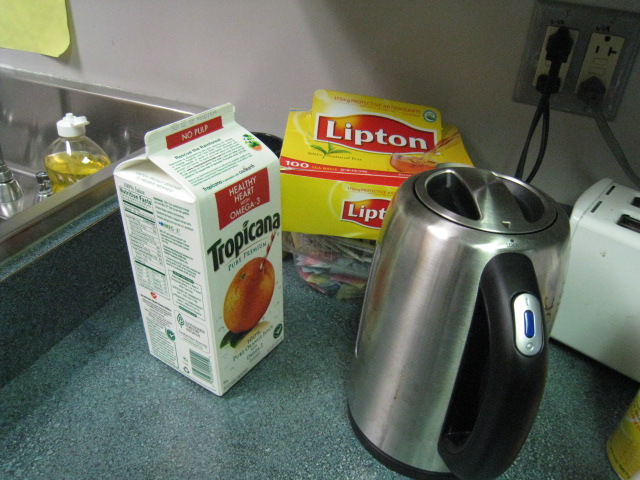}
    \includegraphics[width=0.24 \linewidth]{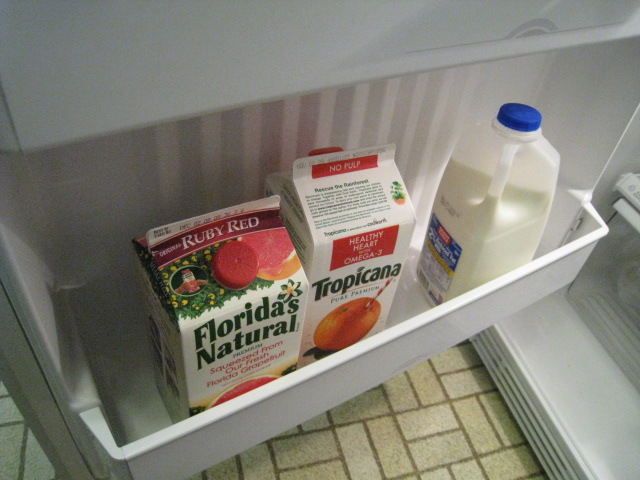}
    \includegraphics[width=0.24 \linewidth]{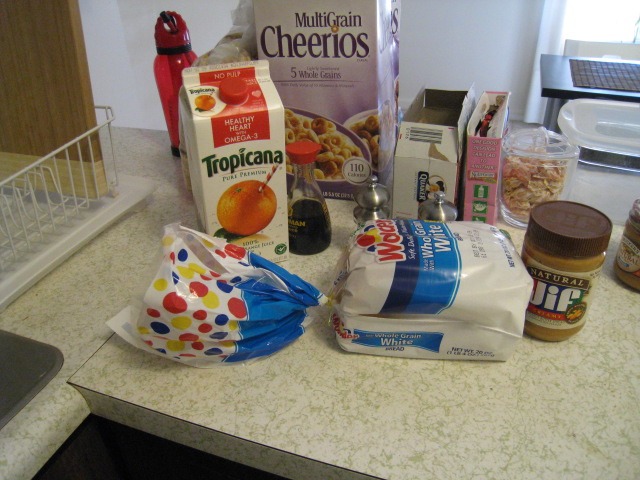} 
    \\
    \caption{A subset of the Orange Juice items in the occluded groceries dataset shown in various background scenes.}
    \label{fig:dataset_example_obj}
\end{figure}

Scene context exists in the common areas/surfaces which hold objects of interest. In many object-detection datasets, this information is not explicitly included, however, identifying the scene types which frequently occur in the dataset can be advantageous. Figure \ref{fig:dataset_example_obj} shows an example of labeled grocery items found in a variety of unlabeled scenes. 

In our methods, scene context is extracted from existing datasets in the form of labels, which is incorporated into the algorithms at various phases during training and pre-evaluation phase. During evaluation, scene-tag prediction is performed using a custom-trained CNN. 

\subsection{Multi-network Fusion (MNF) Algorithm}\label{subsec:multi_network}

\begin{wrapfigure}{r}{0.65\textwidth}
    \centering
    \includegraphics[width=0.99 \linewidth]{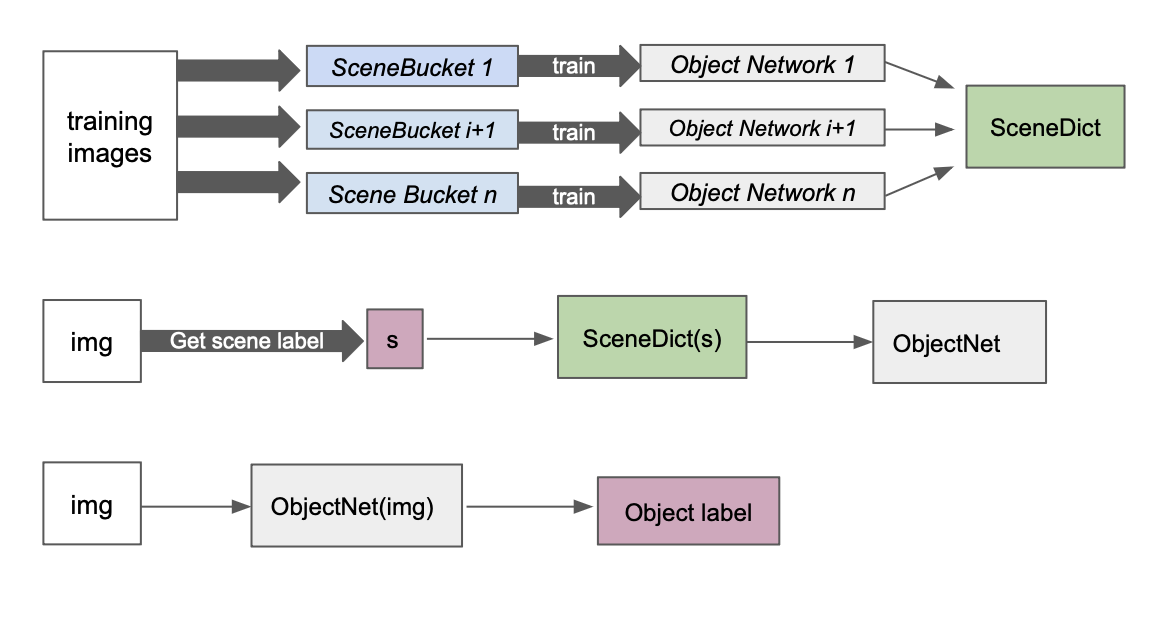}
  \caption{Our Multi-Network Fusion (MNF) Algorithm trains scene-centric object detectors, then chooses the network for each image based on its predicted scene label.}
  \label{fig:methodology_diagram_multinetwork}
\end{wrapfigure}

Our first method selects a scene-specific neural network to detect and to classify each object. 
Each network is custom-trained on a subset of $dataTrain$- those data drawn from scene label $s$, $dataTrain_{s}$. 
Each of these scene networks is stored in a dictionary, $netDict$. Next, each image $i$ in $dataTest$ is mapped by its scene label, $s$ to its custom-trained scene network in $netDict$.
The objects are then classified by the selected network.
The method is further detailed in Algorithm \ref{alg:methodology_multinetwork} and shown in Figure \ref{fig:methodology_diagram_multinetwork}.

To deal with imbalance and ensure the scalability of the algorithm, we ignore all scene types with fewer than $\alpha$ images in the dataset. We set an image-count threshold $\alpha$ to limit the set of labels to those sufficiently represented in the dataset. 

\begin{figure*}
  \centering
  \begin{minipage}{.65\linewidth}
    \begin{algorithm}[H]
    \small
    \caption{\small{Multi-Network Fusion (MNF)}}
    \label{alg:methodology_multinetwork}
    \begin{algorithmic}[1]
    \State \Comment{Training phase of MNF}
    % \State /* Training phase of MNF */
    \For{each scene $s \in Scenes$}
        \State $ dataTrain_{s} \leftarrow \{ i \ for \ i \in dataTrain \ where \ i[s]=s \}$
    \EndFor 
    \State Define $netDict = \{s \ for \ s \in Scenes \}$
    \State $netDict[s] \leftarrow trainNetwork(dataTrain_{s}) \ for \ s \in Scenes$
    \State \Comment{Evaluation phase of MNF}
    \For{each image $i \in dataTest$}
        \State $s_{pred}\leftarrow IdentifyScene(i)$ 
        \State $sceneNet \leftarrow netDict[s_{pred}]$ 
        \State $o_{update} \leftarrow sceneNet(i).label$ 
        \State $P(o_{update}|i) \leftarrow sceneNet(i).score$ 
    \EndFor
    \end{algorithmic}
    \end{algorithm}
  \end{minipage}
\end{figure*}

\subsection{Scene Context Update (SCU) Algorithm}\label{subsec:scene_dist}

This algorithm first extracts the scene label of images and then reclassifies detected objects using object-scene co-occurrence statistics. 

Prior to evaluation, for a given scene label, $s$, and object label, $o$ we find the following probability measures:

\begin{equation}\label{eq:obj_prob}
    P(o) = \frac{1}{m} \sum_{k=1}^{m} I(o, k),
\end{equation}

\begin{equation}\label{eq:scene_prob}
    P(s) = \frac{1}{n} \sum_{h=1}^{n} I(s, h),
\end{equation}

\begin{equation}\label{eq:obj_scene_prob}
    P(o | s ) = \frac{\sum_{k=1}^{m} I(o, s, k) }{ \sum_{k=1}^{m} I(s,k)},
\end{equation}
where $n$ represents the total number of images in the dataset and $h$ the index of an individual image, $m$ represents the total number of object detections over all images and $k$ an individual detection, and where each detection $k$ inherits a scene label from the image in which the detection was made, and
where $I(.)$ is defined as an indicator function which returns 1 if the ground-truth label(s), $s_{gt}$ and/or $o_{gt}$, attached to each detection match the argument(s), shown as $s$ and/or $o$, or 0 otherwise. More specifically, we define three indicator functions of the following form: 
\begin{equation}
I(x,y) = \begin{cases}
1 & \text{if } x \in \ y =x_{gt}, \\
0 & \text{otherwise}, 
\end{cases}
\end{equation}
$for \ (x, y) = \{(o, k), (s, h), ((o, s), k) \}$, representing the object label in each detection, the scene label in each image, and the object-scene pair in each detection, respectively.

\begin{wrapfigure}{r}{0.6\textwidth}
    \centering
    \includegraphics[width=0.99 \linewidth]{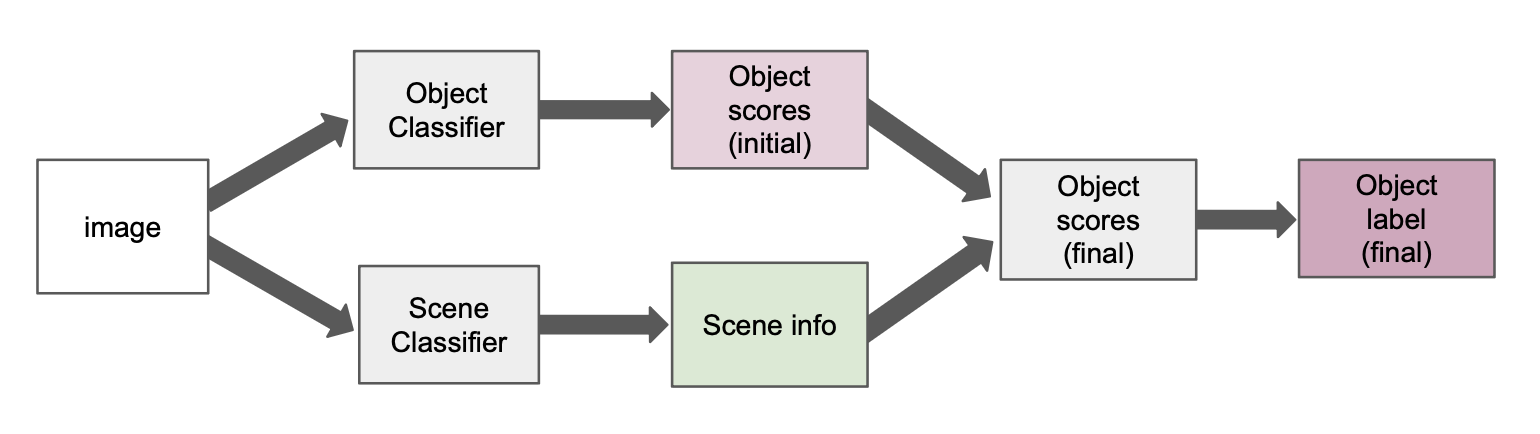}
  \caption{Our Scene-Context Update (SCU) Algorithm updates initial object detections with scene context by incorporating a scene labeler and the dataset statistics.}
  \label{fig:methodology_diagram_scu}
\end{wrapfigure}

These values are used to compute adjusted confidence scores of object predictions output by the original classifier. Given annotated dataset $data$, with subsets reserved for training, $dataTrain$ and testing $dataTest$, the general object-detection \& classification model, $objNet$, is trained using $dataTrain$, and $dataTest$ is first classified using $objNet$. The initial confidence scores for each object detection and classification are saved. These predictions are then fed into our scene-context update algorithm, which incorporates the scene context to refine the original object classification decisions. 

In each image, $i$, for a given bounding box, $bb$, we make use of each potential object score, $P(o_{pot}|bb)$, proposed by $objNet$, for each predicted bounding box. This information is used together with the scene label of $i$, $s$ and their relevant co-occurrence, $P(o_{pot}|s)$ calculated by \eqref{eq:obj_scene_prob}, and shown in Table \ref{table:joint_probs}. The final label, $o_{final}$, is found by: 

\begin{equation}
    o_{final} = \argmax_{o_{pot} \in objPredDict[bb]}P(o_{pot}|bb)P(o_{pot}|s).
\end{equation}

\begin{figure*}
  \centering
  \begin{minipage}{.75\linewidth}
    \begin{algorithm}[H]
    \small
    \caption{\small{Scene-Context Update (SCU)}}
    \label{alg:scene_context_update_2}
    \begin{algorithmic}[1]
     \State \Comment{Training and pre-computation phase of SCU}
    \State $objNet \leftarrow trainNetwork(dataTrain)$
    \For{each $obj, scene \in dataTrain$}
        \State $P(obj|scene) \leftarrow Count(objInScene)$.
    \EndFor
     \State \Comment{Evaluation phase of SCU}
    \For{each image $i \in dataTest$}
        \State $s_{pred} \leftarrow IdentifyScene(i)$
        \For{each predicted bounding box $bb \in i$} 
            \State $objPredDict = \{\}$
            \For {every potential object, $o_{pot}$, from $objNet(bb)$}
                \State $o_{pot} \leftarrow o_{pot}.label$ 
                \State $P(o_{pot}|bb) \leftarrow o_{pot}.score$ 
                \State $objPredDict.insert(\{o: P(o_{pot}|bb)\})$
            \EndFor
            \State $o_{final} = \argmax_{o_{pot} \in objPredDict[bb]}P(o_{pot}|bb)P(o_{pot}|s_{pred})$ 
        \EndFor
    \EndFor
    \end{algorithmic}
    \end{algorithm}
  \end{minipage}
\end{figure*}

Note $s$ refers to the predicted scene label attached to each image which contains $bb$. This update rule is applied for each feasible bounding box, $bb$, to obtain a new prediction with label $o_{updated}$ and score $P(o_{updated})$. The rule reflects an iteration over each potential object, $o_{pot}$ where $P(o_{pot}|bb)$ denotes the probability score predicted by $objNet$ and $P(o_{pot}|s)$ denotes the relevant value in the object-scene co-occurrence table, which may be pre-computed from the training set using \ref{eq:obj_scene_prob}. Note that if the object scores for a given bounding box, $bb$, are sufficiently close to zero, they are not included in the results, so not every $o \in O$ is considered, but rather every $o_{pot} \in objDict[bb]$, the set of predicted objects, where $objDict[bb]  \subseteq O$. 
The $P(O|bb)$ for $o \notin {O_{pot}}$ are therefore set to zero in the algorithm for convenience. This process is further illustrated in Algorithm \ref{alg:scene_context_update_2} and Figure \ref{fig:methodology_diagram_scu}.

\section{Experiments}\label{sec:experiments}

\subsection{Datasets}\label{sec:dataset}
The Occluded Groceries dataset\cite{hsiao2010making} features ten unique grocery items. As each of the occluded objects are photographed in several distinct locations around a single home, the dataset offers rich background information that can potentially be leveraged as scene context. It offers two distinct datasets: the unoccluded dataset features the objects photographed in isolation, containing 250 images, the occluded dataset features the objects in natural settings which feature heavy occlusions, such as partial coverage by other objects, shadows, or variable lighting, containing 500 images with 750 object detections. Figure \ref{fig:CACN_groceries_examples} in Supplementary Materials shows an example object from both sets for comparison. Figure \ref{fig:dataset_example_obj} shows a subset of the object label Orange Juice occurring in various scenes and featuring various occlusions. Table \ref{table:joint_probs} shows the object-scene co-occurrences of each object-scene pair in the dataset, formulated as $P(O|S)$. 

The TACO dataset \cite{taco2020} is an open-source state-of-the-art dataset for litter detection in the wild featuring over 2,000 annotated images. Objects are of various sizes and feature various occlusions in their natural environments such as partial occlusions caused by dirt or grass, fence chains or other pieces of litter. Each object is labeled into category and supercategory and also includes a label of the background, or scene, surrounding it. Supplementary Materials Figure \ref{fig:taco_examples} shows examples of the types of scenes and occlusions present in the TACO dataset. Table \ref{table:joint_probs_taco} shows the object-scene co-occurrences of a subset of the dataset, featuring the three most common scenes and the 10 most common objects, formulated as $P(O|S)$. 

These two datasets demonstrate real-world use cases, such as a smart refrigerator or litter-collecting robot. We acknowledge that these datasets are small or not specifically designed for occlusion testing, however, given the need for scene context in the experiments, they were the most suitable which satisfied both criteria. Current SOTA datasets for occlusion handling do not include objects captured in natural environments \cite{wu2023robustness} \cite{wang2020robust}.

\begin{minipage}[b]{.45\textwidth}
  \scriptsize
  \centering
  \begin{tabular}{l c c c}
    \hline
    & Cupboard  & Counter & Refrig.  \\
    \midrule
    Can Chowder & 20.6\% & 16.8\% & 0\%  \\
    Can Soymilk & 0\% & 3.8\% & 20\% \\
    Can Tomatosoup & 25.8\% & 12.3\% & 0\%  \\
    Carton OJ & 0\% & 3.8\% & 21\%  \\
    Carton Soymilk & 0\% & 0\% & 23.8\%  \\
    Diet Coke & 0\% & 8.4\% & 12.8\% \\
    HC Potroast & 20.6\% & 9.7\% & 2.8\%  \\
    Juicebox & 0\% & 7.1\% & 18.5\%  \\
    Rice Tuscan & 16.3\% & 20.1\% & 0\%  \\
    Rice Pilaf & 16.3\% & 17.5\% & 0\%  \\
    \bottomrule
    \\
  \end{tabular}
  \captionof{table}{Object-scene co-occurrences in Occluded Groceries dataset, formulated as $P(O|S)$.}
  \label{table:joint_probs}
\end{minipage}\qquad
\begin{minipage}[b]{.45\textwidth}
  \scriptsize
  \centering
  \begin{tabular}{l c c c}
    \hline
    & Pavement & Vegetation & Sand... \\
    \midrule
        Plastic bag.. & 24.9\% & 31.8\% & 29.7\%\\
        Cigarette & 27.0\% & 12.2\% & 17.7\% \\
        Bottle & 09.7\% & 14.2\% & 11.0\%  \\
        Bottle cap & 07.5\% & 08.8\% & 09.3\%  \\
        Carton & 06.9\% & 06.8\% & 06.2\% \\
        Can & 05.2\% & 08.3\% & 04.9\% \\
        Cup & 05.4\% & 05.6\% & 05.7\% \\
        Straw & 05.6\% & 04.6\% & 06.5\% \\
        Paper & 04.7\% & 03.9\% & 04.0\%  \\
        Styrofoam.. & 03.1\% & 03.7\% & 04.9\%  \\
        \bottomrule
        \\
      \end{tabular}
  \captionof{table}{Object-scene co-occurrences of TACO dataset, formulated as $P(O|S)$.}
  \label{table:joint_probs_taco}
\end{minipage}

\subsection{Models and Baselines}
In all experiments, a custom-trained Resnet-18 model (pre-trained on ImageNet) is used for scene detection. For object detection, we use a custom-trained YOLO (v3) \cite{redmon2016you} network, which is included as a baseline comparison. All object-detection models are trained on NVIDIA GeForce GTX 1080 Ti for 6000 steps, with batch size of 64, with 16 subdivisions, and a learning rate of 0.001.

\begin{wraptable}{r}{0.65\textwidth}
  \scriptsize
  \centering
    \begin{tabular}{l l l l l l l}
    \toprule
    Object & YOLO  & & MNF  & & SCU  \\
    - & Pr. & Re.  & Pr.  &Re.  & Pr.  & Re. \\
    \midrule
        Can Chowder 
            & 71.4\% & 68.2\% 
            & \textcolor{blue}{\textbf{100\%}} &  \textcolor{blue}{\textbf{82.4\%}}    
            & \textcolor{red}{68.0\%} &  \textcolor{blue}{77.3\%} \\
        Can Soymilk 
            & 57.1\% & \textbf{88.9\%} 
            & \textcolor{blue}{\textbf{88.9\%}} & \textcolor{red}{\textbf{80.0\%}} 
            & 57.1\% &  \textbf{88.9\%} \\
        Can Tomatosoup 
            & \textbf{100\%} & 53.8\% 
            & \textbf{100\%} &  \textcolor{blue}{\textbf{85.7\%}} 
            & \textbf{100\%} &  53.8\% \\
        Carton OJ 
            & \textbf{72.7\%} &  53.3\% 
            & \textcolor{red}{62.5\%} &  \textcolor{blue}{\textbf{83.3\%}}  
            & \textcolor{red}{61.5\%} & 53.3\% \\
        Carton Soymilk 
            & 86.7\% &  81.3\% 
            & \textcolor{blue}{\textbf{88.2\%}} & \textcolor{blue}{\textbf{88.2\%}}   
            & \textcolor{red}{81.3\%} &  81.3\% \\
        Diet Coke 
            & 54.5\% &  \textbf{100\%} 
            & \textcolor{blue}{\textbf{77.8\%}} &  \textbf{100\%}  
            & 54.5\% &   \textbf{100\%} \\
        HC Potroast 
            & \textbf{80.0\%} & \textbf{100\%} 
            & \textcolor{red}{75.0\%} &  \textcolor{red}{66.7\%}   
            & \textcolor{red}{77.8\%} &  \textcolor{red}{87.5\%} \\
        Juicebox 
            & 90.0\% & 56.3\% 
            & \textcolor{red}{85.7\%} & \textcolor{blue}{\textbf{80.0\%}}   
            & \textcolor{blue}{\textbf{100\%}} & 56.3\% \\
        Rice Tuscan 
            & 75.0\% &   69.2\% 
            & \textcolor{blue}{\textbf{100\%}} & \textcolor{blue}{\textbf{94.1\%}} 
            & \textcolor{blue}{81.8\%} &  69.2\% \\
        Rice Pilaf 
            & 33.3\% &   71.4\% 
            & \textcolor{blue}{\textbf{75.0\%}} &  \textcolor{blue}{\textbf{100\%}} 
            & \textcolor{blue}{41.7\%} &  71.4\% \\
        Total 
            & 70.3\% &   70.3\% 
            & \textcolor{blue}{\textbf{85.5\%}} &  \textcolor{blue}{\textbf{85.5\%}}
            & \textcolor{blue}{71.0\%} &  \textcolor{blue}{71.0\%} \\
    \bottomrule
    \\ 
  \end{tabular}
\caption{Classification performance of our Multi-Network Fusion (MNF) Algorithm on occluded groceries dataset shown against YOLO as a baseline method. Weights are trained only on occluded images and tested only on occluded images.}\label{table:occluded_weights}
\end{wraptable} 

We perform three experiments to test our context algorithms, each featuring object-detection models custom-trained under differing degrees of occlusion. The first experiment is run on both the Occluded Groceries and TACO datasets, while the second and third are run on the Occluded Groceries dataset only. The models in the first experiment are trained on occluded images only, which allows us to test both the Multi-Network Fusion (MNF) algorithm, which depends on the greater scene context provided in the occluded images at training time, and the Scene-Context Update (SCU) algorithm, which can work with any type of training images.  The models in the second experiment are trained on unoccluded images only, which prohibits the use of the MNF algorithm, but allows for implementation of SCU and an additional baseline model for comparison, Contextually-Aware Compositional Nets (CACN)\cite{kortylewski2021compositional}, which by design should be trained on unoccluded images. Lastly, the models in the third experiment are trained on a combination of occluded and unoccluded images, and include the SCU against the baseline YOLO model.

\begin{figure}[!h]
    \includegraphics[width=0.32\linewidth]{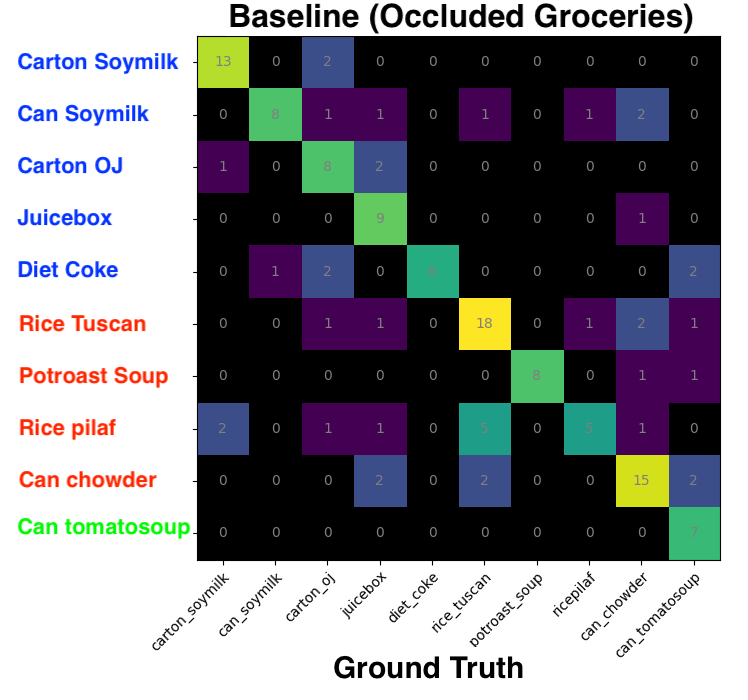}
    \includegraphics[width=0.32\linewidth]{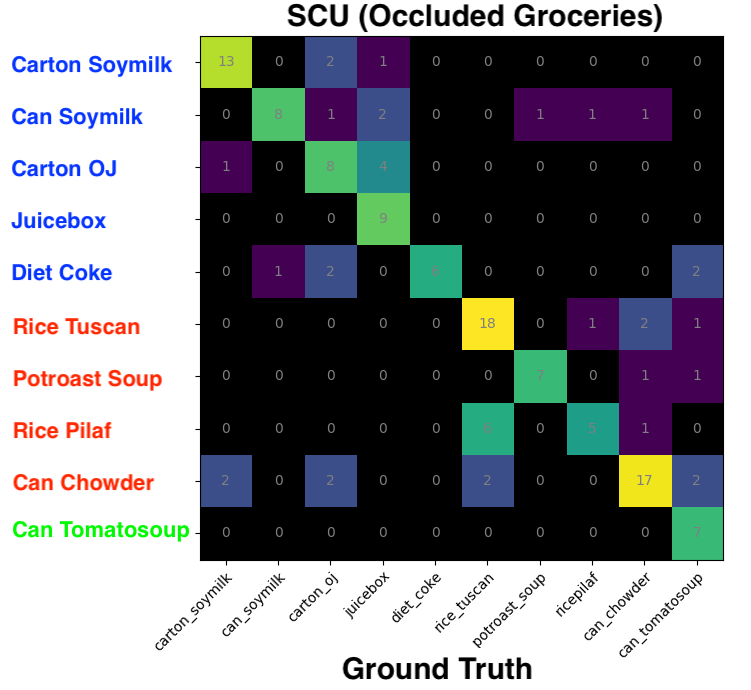}
    \includegraphics[width=0.32\linewidth]{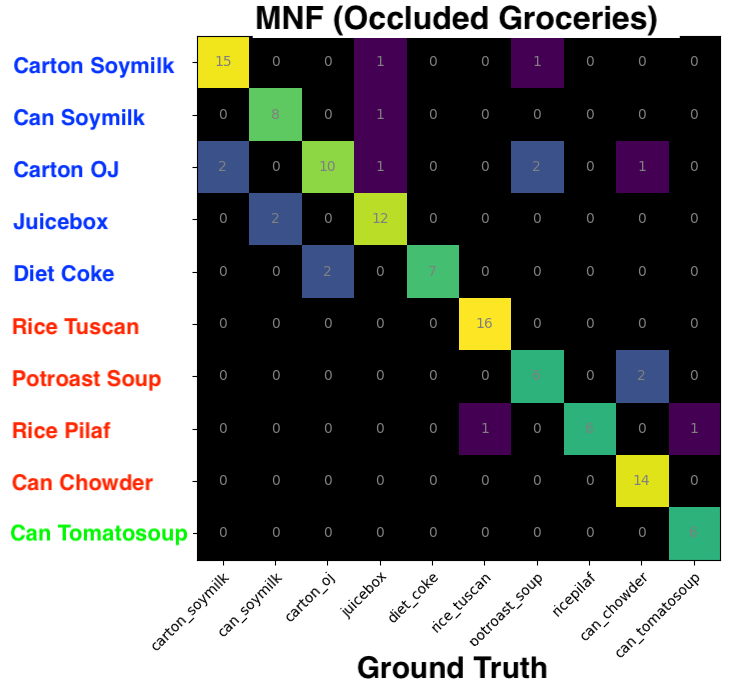}
    \\
\caption{Confusion matrices produced from models trained on the Occluded Groceries dataset, corresponding to Table \ref{table:occluded_weights}.}
\label{fig:confusion_matrix_occluded}
\end{figure}

\begin{figure}[!h]
    \includegraphics[width=0.32\linewidth]{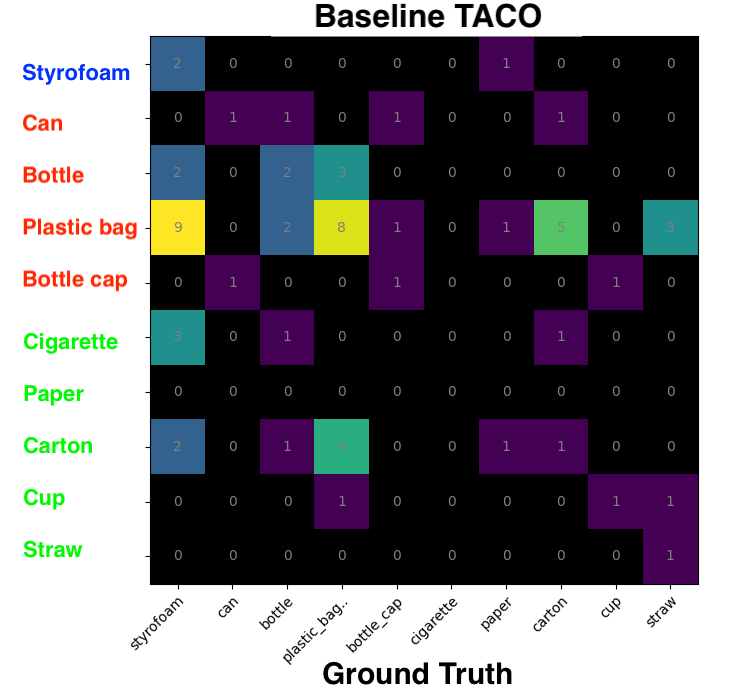}
    \includegraphics[width=0.32\linewidth]{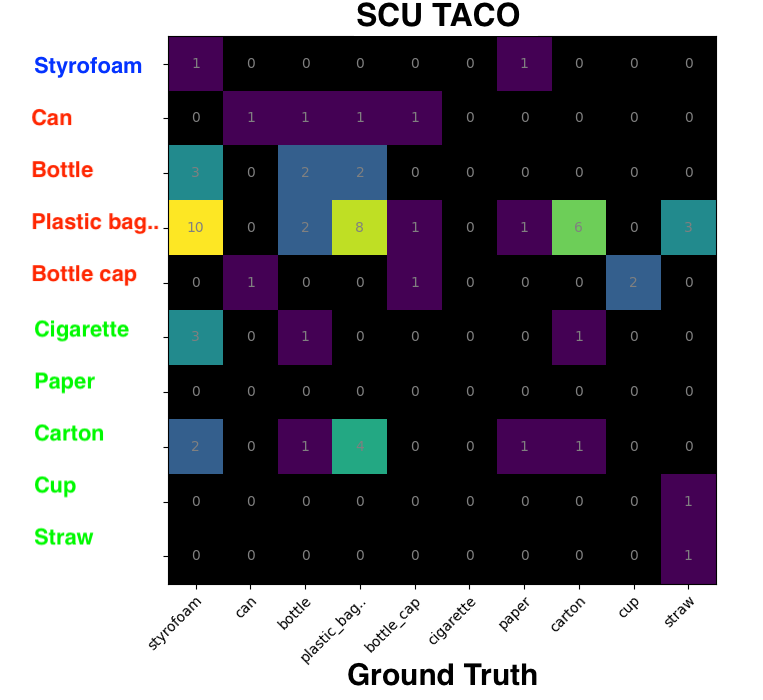}
    \includegraphics[width=0.32\linewidth]{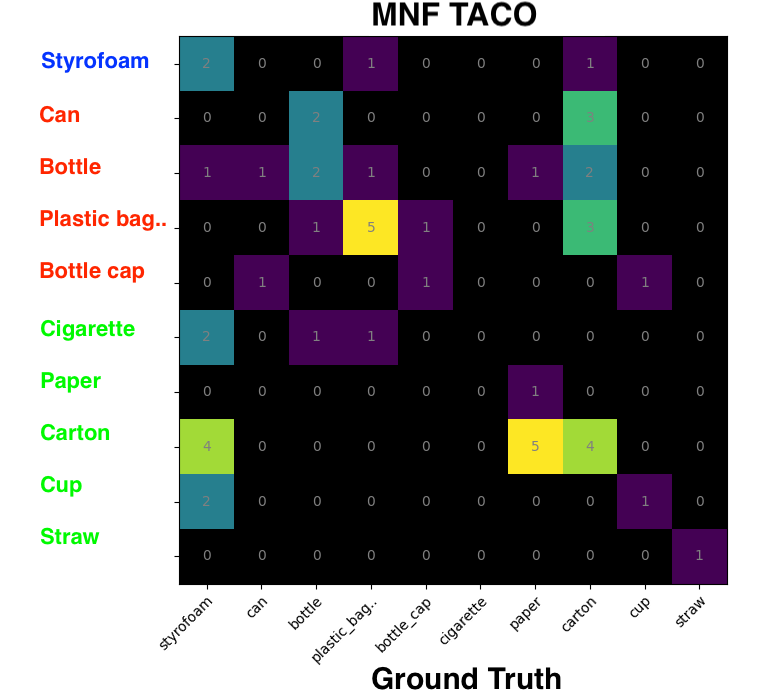}
    \\
\caption{Confusion matrices produced from experiments on the TACO dataset and corresponding to Table \ref{table:taco_results}.}
\label{fig:confusion_matrix_TACO}
\end{figure}

\subsection{Results and Discussion}

Results shown from all experiments include the precision and recall for each of the featured objects. Changes in performance for both algorithms are denoted by color, with blue denoting a performance increase, and red a decrease, against the standard baseline model for each object category, and the highest score(s) are shown in bold. Precision-Recall curves are also included in Supplementary Material \ref{app:pr_curves}. Additionally, confusion matrices for each of the algorithms, corresponding to the experiments shown in Results Table \ref{table:occluded_weights} tested on the Occluded Groceries dataset, and Results Table \ref{table:taco_results} tested on the TACO dataset, are respectively shown in Figures \ref{fig:confusion_matrix_occluded} and \ref{fig:confusion_matrix_TACO}, and confusion matrices from subsequent experiments are included in Supplementary Material \ref{app:confusion_matrices}. Confusion Matrices are colored in log-scale with yellow denoting the highest number of detections, purple the lowest, and black, zero.

\begin{wraptable}{r}{0.65\textwidth}
  \scriptsize
  \centering
    \begin{tabular}{l l l l l l l}
    \toprule
    Object & YOLO  & & MNF  & & SCU  \\
    - & Pr. & Re.  & Pr.  &Re.  & Pr.  & Re. \\
    \midrule
        Plastic bag..
            & 32.6\% & 51.9\% 
            & \textbf{\textcolor{blue}{62.5\%}} & \textbf{\textcolor{blue}{58.8\%}} 
            & 32.6\% & \textcolor{blue}{55.6\%} \\
        Cigarette
            & 00.0\% & 00.0\%  
            & 00.0\% & 00.0\% 
            & 00.0\% & 00.0\%  \\
        Bottle
            & 20.0\% & 25.0\% 
            & \textbf{\textcolor{blue}{25.0\%}} & \textbf{\textcolor{blue}{33.3\%}}
            & \textcolor{blue}{22.2\%} & 25.0\%    \\
        Bottle cap 
            & 25.0\% & 33.3\% 
            & \textbf{\textcolor{blue}{33.3\%}} & \textbf{\textcolor{blue}{50.0\%} }
            & \textcolor{red}{20.0\%} & 33.3\%  \\
        Carton
            & 30.8\% & 22.2\% 
            & \textbf{\textcolor{blue}{41.2\%}} & \textbf{\textcolor{blue}{38.9\% }}
            & \textcolor{red}{28.6\%} & 22.2\%  \\ 
        Can
            & \textbf{25.0\%} & \textbf{50.0\% }
            & \textcolor{red}{00.0\%} & \textcolor{red}{00.0\%}
            & \textbf{25.0\%} & \textbf{50.0\% } \\
        Cup
            & \textbf{25.0\%} & 25.0\% 
            & \textbf{25.0\%} & \textbf{\textcolor{blue}{50.0\%}}
            & \textcolor{red}{00.0\%} & \textcolor{red}{00.0\%}  \\
        Straw 
           & \textbf{100\%} & 25.0\% 
           & \textcolor{red}{50.0\%} & \textbf{\textcolor{blue}{100\%}}  
           & \textbf{100\%} & 25.0\%  \\
        Paper
            & 00.0\% & 00.0\% 
            & \textbf{\textcolor{blue}{100\%}} & \textbf{\textcolor{blue}{14.3\% }} 
            & 00.0\% & 00.0\%  \\
        Styrofoam..
            & \textbf{66.7\%} & 11.1\% 
            & \textcolor{red}{40.0\%} & \textbf{\textcolor{blue}{18.2\%}} 
            & \textcolor{red}{59.1\%} & \textcolor{red}{05.6\% } \\
        Total 
            & 29.7\% & 29.7\% 
            & \textbf{\textcolor{blue}{37.9\%}} & \textbf{\textcolor{blue}{37.9\%}}
            & \textcolor{red}{28.6\%} & \textcolor{red}{28.6\%}  \\
    \bottomrule
    \\ 
  \end{tabular}
\caption{Classification performance of SCU and MNF algorithm shown against baseline YOLO model. Weights are tested and trained on a subset of the TACO litter detection dataset.}\label{table:taco_results}
\end{wraptable} 

Table \ref{table:occluded_weights} highlights the performance gains of the two algorithms. MNF shows a substantial improvement over the baseline in average recall and precision, while SCU also shows a minor improvement.  Interestingly, the two algorithms show different increases and decreases with respect to the baseline at the individual object categories. This could imply that one scene-context algorithm could be more desirable than the other in certain domains and use-cases. To further support this claim, the corresponding confusion matrices of the three models are shown in Figure \ref{fig:confusion_matrix_occluded}, with object labels in matrix rows and columns clustered by scene, as further detailed in Supplementary Materials \ref{app:confusion_matrices}. The range of errors is markedly decreased in both scene-context algorithms from the baseline, with MNF showing the lowest range of errors. Both algorithms show a tendency to confuse one object with another object in the same scene, while the confusions by the baseline are less biased by the scene, which can be observed by the prominence of black regions in the lower-left and upper right corners.  A similar trend is observed in the experiments on the TACO dataset, shown in Results Table \ref{table:taco_results}, and its related confusion matrices, shown in Figure \ref{fig:confusion_matrix_TACO}, although slightly less pronounced, and with the diagonal trend line, indicating true positives, appearing the brightest.

\subsection{Alternate Training Methodologies}\label{subsec:alternate_training}

\begin{wraptable}{r}{0.65\textwidth}
  \scriptsize
  \centering
  \begin{tabular}{l l l l l l l}
    \toprule
    Object  & YOLO & & SCU & & CACN   \\
    - & Pr. & Re. & Pr. & Re.  & Pr. & Re. \\
    \midrule
        Can Chowder 
            & \textbf{45.7\%} & 62.7\% 
            & \textcolor{red}{36.9\%}  & \textcolor{blue}{\textbf{80.4\%}} 
            & 04.0\% &  06.5\% \\
        Can Soymilk 
            & \textbf{42.0\%} & 60.3\% 
            & \textcolor{blue}{\textbf{43.1\%}} & \textbf{60.3\%}
            & 04.0\% & 06.9\% \\
        Can Tomatosoup 
            & 67.7\% & 48.8\%
            & \textcolor{blue}{\textbf{71.9\%}} & \textcolor{blue}{\textbf{53.5\%}} 
            & 06.0\% & 07.9\% \\
        Carton OJ 
            & 50.0\% & \textbf{53.4\%} 
            & \textcolor{blue}{\textbf{66.3\%}} &  \textbf{53.4\%} 
            & 02.0\% & 09.1\% \\
        Carton Soymilk 
            &44.5\% & 61.6\%
            & \textcolor{blue}{\textbf{69.8\%}} & \textcolor{blue}{\textbf{67.7\%}} 
            & 04.0\% & 06.1\% \\
        Diet Coke 
            & 60.0\% & 44.6\% 
            & \textcolor{blue}{\textbf{64.8\%}} & \textcolor{blue}{\textbf{47.3\%}} 
            & \textcolor{purple}{18.0\%} & \textcolor{purple}{12.7\%} \\
        HC Potroast 
            & 48.4\% & 62.5\% 
            & \textcolor{blue}{\textbf{57.4\%}} & \textcolor{blue}{\textbf{91.1\%}}  
            & 08.0\% & 07.5\% \\
        Juicebox 
            & \textbf{50.0\%} & 13.2\% 
            & \textcolor{red}{43.7\%} & \textcolor{blue}{\textbf{18.4\%}} 
            & 08.0\% & 10.0\% \\
        Rice Tuscan 
            & \textbf{73.2\%} &  47.7\% 
            & \textcolor{red}{63.5\%} & \textcolor{blue}{\textbf{62.8\%}} 
            & \textcolor{purple}{38.0\%} & \textcolor{purple}{11.5\%} \\
        Rice Pilaf 
            & 55.1\% &  43.4\% 
            & \textcolor{blue}{\textbf{65.4\%}} & \textcolor{blue}{\textbf{53.5\%}} 
            & \textcolor{purple}{06.0\%} & \textcolor{purple}{03.8\%} \\
        Total 
            & 51.1\% & 51.1\% 
            & \textcolor{blue}{\textbf{57.4\%}} & \textcolor{blue}{\textbf{57.4\%}} 
            & 09.8\% & 08.9\% \\
    \bottomrule
    \\
  \end{tabular}
  \caption{Classification performance of our Scene-Context Update (SCU) Algorithm on occluded groceries dataset shown against YOLO and Context Aware Compositional Nets (CACN) as baseline comparison methods. YOLO weights and CACN model are trained only on unoccluded images and tested on occluded images. }\label{table:unoccluded_weights}
\end{wraptable}

Table \ref{table:unoccluded_weights} shows performance gains of SCU when trained on unoccluded images only. The algorithm shows an increase of $>6\%$ in average performance as well as in the majority of the individual categories. The lower overall performance, compared to Table \ref{table:occluded_weights}, is expected given the absence of occlusions in the training dataset and the heavy occlusions in the test set. Given the absence of scene context in the training set, MNF is excluded from the experiment, but an additional baseline comparison model is included. This model shows significantly lower performance than both the SCU algorithm as well as the YOLO baseline, however, we chose to still include it due to a lack of comparable baseline models. These low scores can be attributed to its ability to work on partially occluded images, but perhaps not those in crowded scenes. The dataset we have tested it on differs substantially from those which it was originally tested on, which feature single items with partial occlusions. Given that our datasets feature natural occlusions in crowded scenes, we believe this can explain the substantially lower performance scores. Further comparison of these two datasets are included in Supplementary Material \ref{app:baseline_comparisons}. 

\begin{wraptable}{r}{0.5\textwidth}
  \scriptsize
  \centering
  \begin{tabular}{l l l l l}
    \toprule
    Object & YOLO & & SCU &  \\
    - & Pr. & Re.  &  Pr. & Re.  \\
    \midrule
        Can Chowder & \textbf{74.3\%}  & 81.3\% 
            & \textcolor{red}{66.0\%} & \textcolor{blue}{\textbf{96.9\%}}\\
        Can Soymilk & 67.9\% & \textbf{73.1\%} 
            & \textcolor{blue}{\textbf{73.1\%}} & \textbf{73.1\%}  \\
        Can Tomatosoup & 68.4\%  & \textbf{86.7\%}
            & \textcolor{blue}{\textbf{72.2\%}} & \textbf{86.7\%}   \\
        Carton OJ & \textbf{86.4\%}  &  \textbf{84.4\%} 
            & \textbf{86.4\%}  &\textbf{84.4\%} \\
        Carton Soymilk & 69.6\%   & \textbf{90.7\%}
            & \textcolor{blue}{\textbf{78\%}}  & \textbf{90.7\%}  \\
        Diet Coke & \textbf{83.0\%}  & 69.6\%  
            & \textcolor{red}{80.0\%}  & \textcolor{blue}{\textbf{71.4\%}}   \\
        HC Potroast & \textbf{96.4\%}  & \textbf{96.4\%} 
            & \textbf{96.4\%}   & \textbf{96.4\%} \\
        Juicebox & 92.6\%   &  61.0\% 
            & \textcolor{blue}{\textbf{100\%}}  & \textcolor{blue}{\textbf{63.4\%}}   \\ 
        Rice Tuscan & 78.1\%  & \textbf{75.8\%}  
            & \textcolor{blue}{\textbf{86.2\%}}  & \textbf{75.8\%}\\
        Rice Pilaf & 76.9\%   & 83.3\% 
            & \textcolor{blue}{\textbf{81.9\%}}  &  \textbf{83.3\%}   \\
        Total  & 79.2\%   & 79.2\%  
            &  \textcolor{blue}{\textbf{81.1\%}}   & \textcolor{blue}{\textbf{81.1\%}}   \\
    \bottomrule
    \\
  \end{tabular}
  \caption{Classification performance of our Scene-Context Update (SCU) Algorithm on occluded groceries dataset shown against YOLO as a baseline. YOLO weights are trained on both occluded and unoccluded images, and tested only on occluded images.}\label{table:unoccluded_occluded_weights}
\end{wraptable}

Strongest classification performances overall were observed when training on both occluded and unoccluded images, as shown in Table \ref{table:unoccluded_occluded_weights}. However, improvement of SCU over baseline is most pronounced when training on unoccluded objects alone. Overall, the SCU algorithm shows an improvement over baselines, having the large majority of the best performance scores over the individual categories in both tables. The increase between the baseline in Table \ref{table:unoccluded_occluded_weights} is smaller, at $1.9\%$, while that of Table \ref{table:unoccluded_weights} is larger at $6.3\%$. The larger performance increase against the baseline by the weights trained on the unoccluded images supports that the SCU method would offer additional robustness, given its ability to perform well when tested on images featuring occlusions which were entirely absent during the training process. Confusion matrices included in Supplementary Materials \ref{app:confusion_matrices} demonstrate a similar pattern across all experiments of making less out-of-scene errors, showing the advantage of scene context in object detection. 

\section{Conclusions and Future Directions}
In this study, we presented two novel algorithms which fuse scene context to improve the classification of occluded objects. We tested our algorithms on two challenging datasets featuring occluded objects. Results of the two algorithms demonstrated a notable performance increase. Additionally, we explored alternate training methodologies for improved occlusion handling.

As aforementioned, finding suitable baseline comparisons and datasets was a challenge, pointing to the need for more diverse and realistic occlusion-handling datasets. Other steps forward include further investigation into the types of errors shown in confusion matrices, to incorporate error-handling techniques into our design. As the errors seem to exist between very similar objects, a new context-based algorithm could be developed to resolve this specific problem. 

\section{Acknowledgments}
C. King thanks Fordham GSAS for continued funding and support and the opportunity to present the early stages of this work at GSAS Research Day 2024, as well as D. Fouhey for the opportunity to present this work at NYC Vision Day 2025. She is grateful for support under the Maguire-Noz fellowship. In addition to her co-authors, she expresses gratitude to F. Hsu, E. Aminoff, A. Alfatemi, N. Paykari and E. Roginek for useful conversations and feedback, as well as to A. Milekhin for moral support.

\bibliography{egbib}
\newpage
\section{Supplementary Materials}

\subsection{Baseline Comparison}\label{app:baseline_comparisons}
\begin{figure}[h!]
    \includegraphics[width=0.24\linewidth]{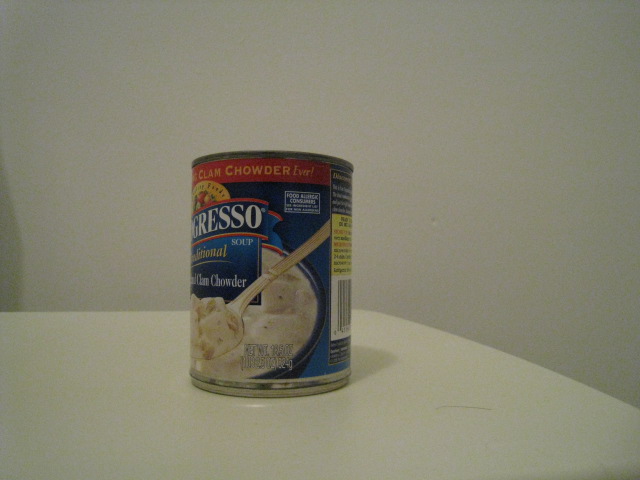}
    \includegraphics[width=0.24\linewidth]{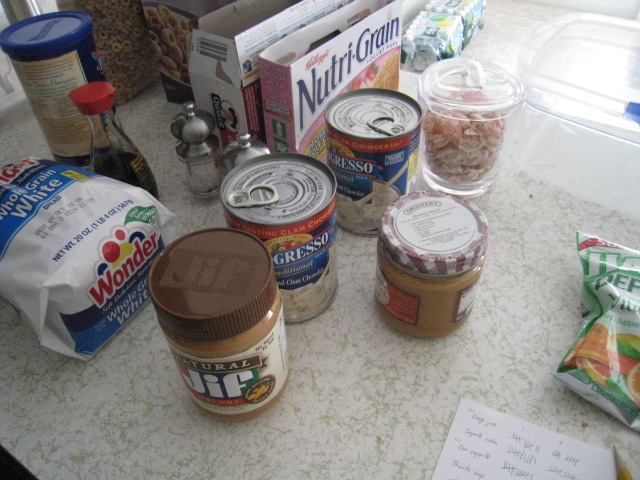}
    \includegraphics[width=0.24\linewidth]{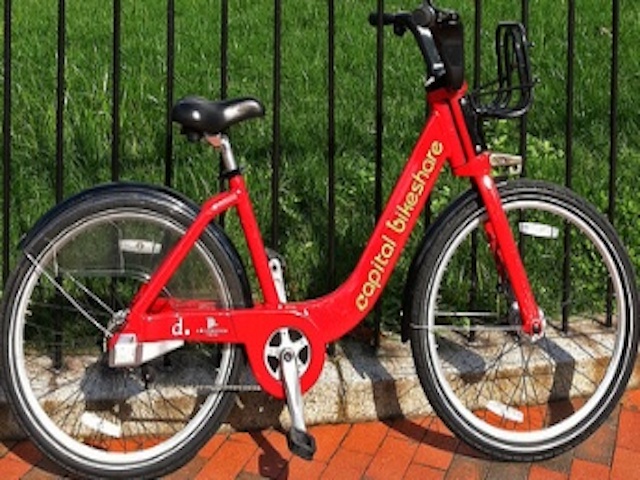}
    \includegraphics[width=0.24\linewidth]{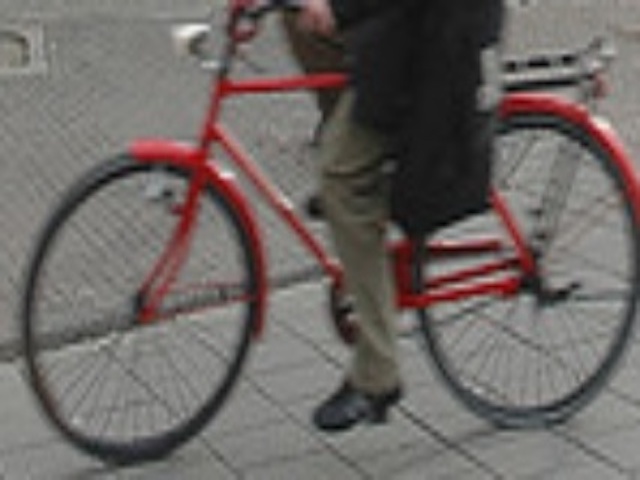}
    \caption{Left half: The Occluded Groceries dataset features two subsets containing the 10 objects. Shown is an example of the Can Chowder object found in the unoccluded set (left) and the occluded set (right). Right half: The CACN model is trained on subsets of COCO.  Shown is an example of a Bicycle found in the unoccluded set (left) and the occluded set (right).}
    \label{fig:CACN_groceries_examples}
\end{figure}

The CACN baseline was trained and tested on a different dataset to show its ability to handle partial occlusions on individual items. The image set used for both testing and training was perhaps a bit less challenging than the one used in this paper, which may explain the significant drop in performance scores when using the occluded groceries dataset. An example of a zero-occlusion image for training and a level-one occlusion image for testing from the original paper are shown in Figure \ref{fig:CACN_groceries_examples} on the left, which are quite different from those used in this paper, shown on the right for comparison. 

\subsection{Datasets}\label{app:datasets}
\begin{figure}[h!]
    \includegraphics[width=0.24\linewidth]{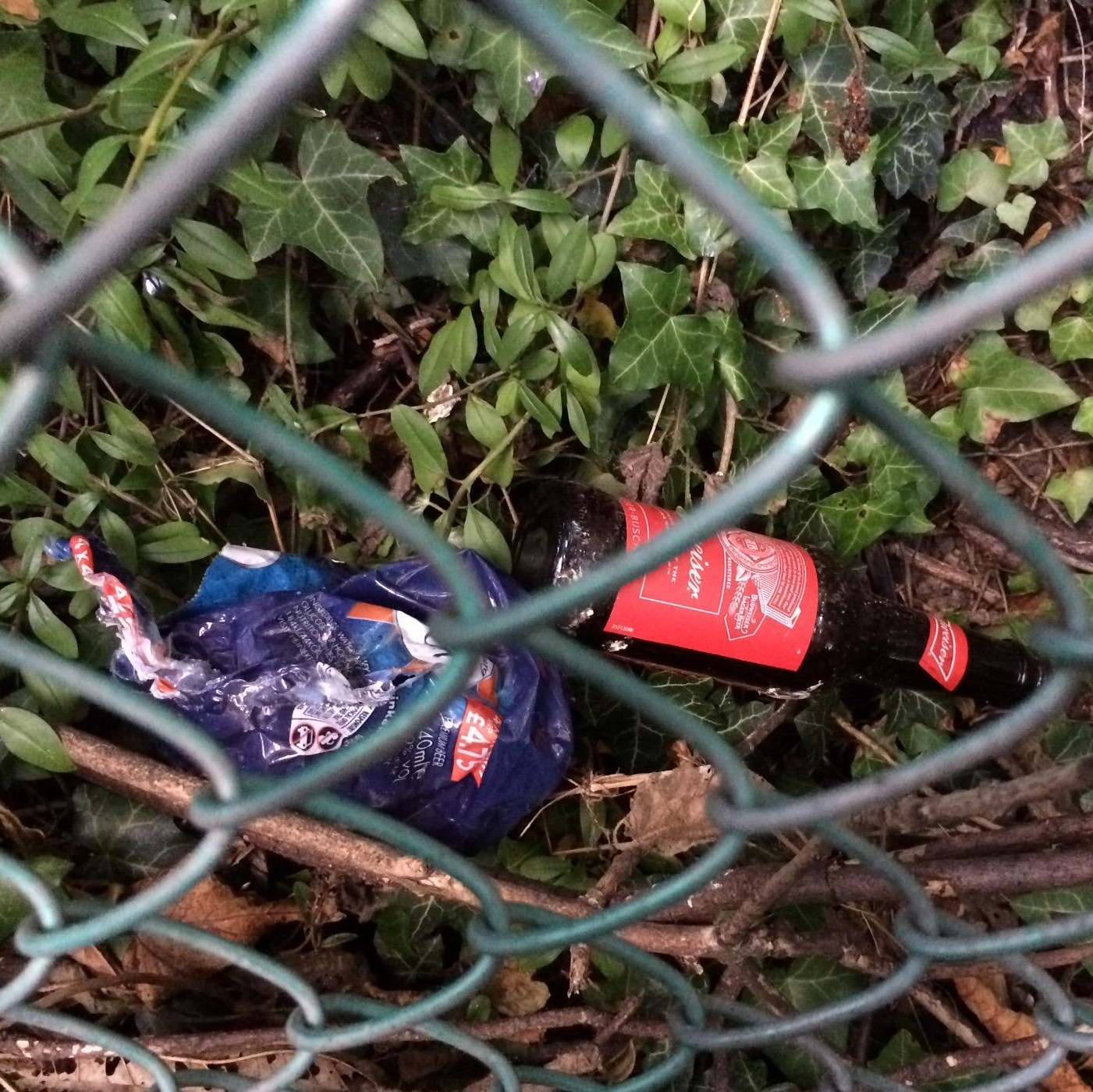}
    \includegraphics[width=0.24\linewidth]{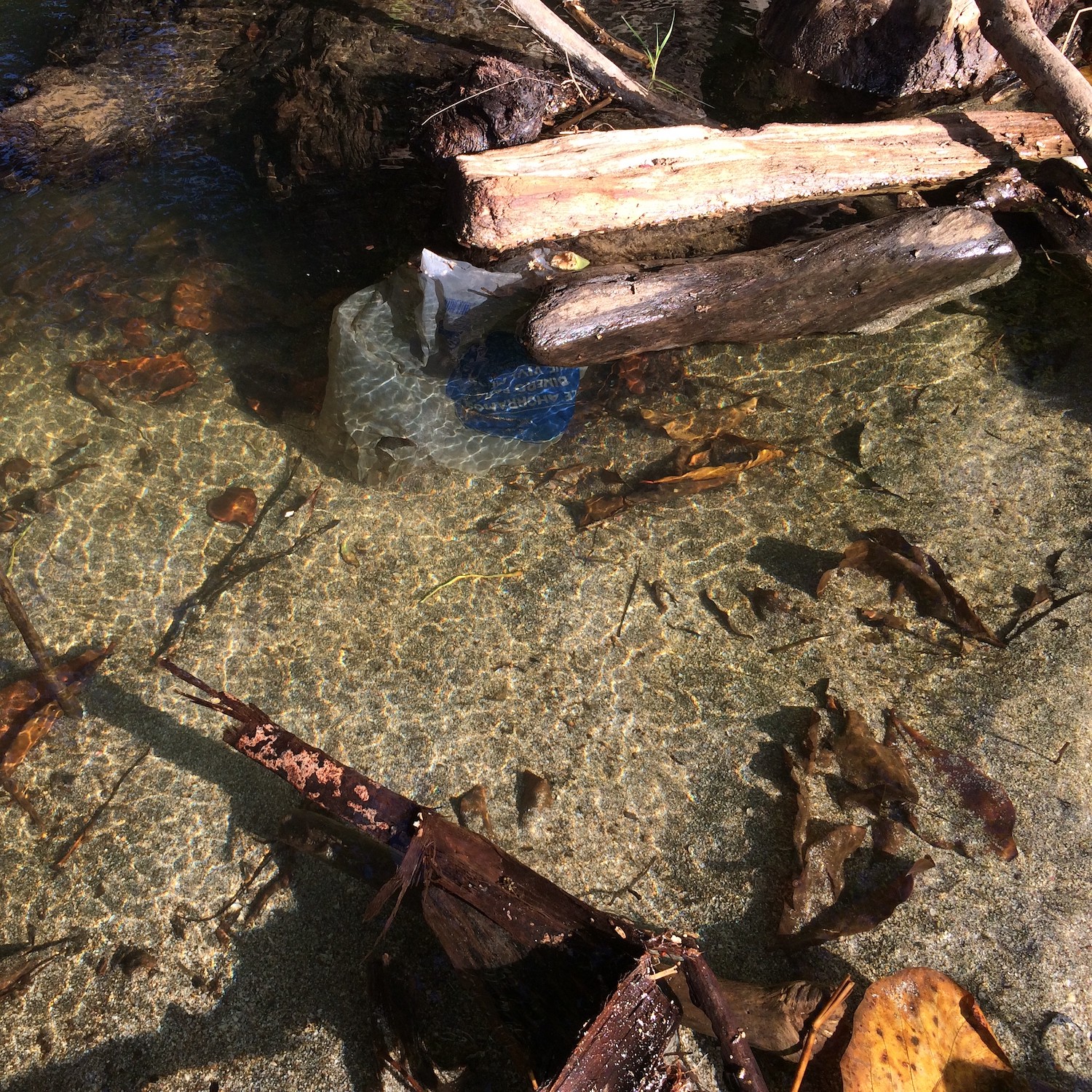}
    \includegraphics[width=0.24\linewidth]{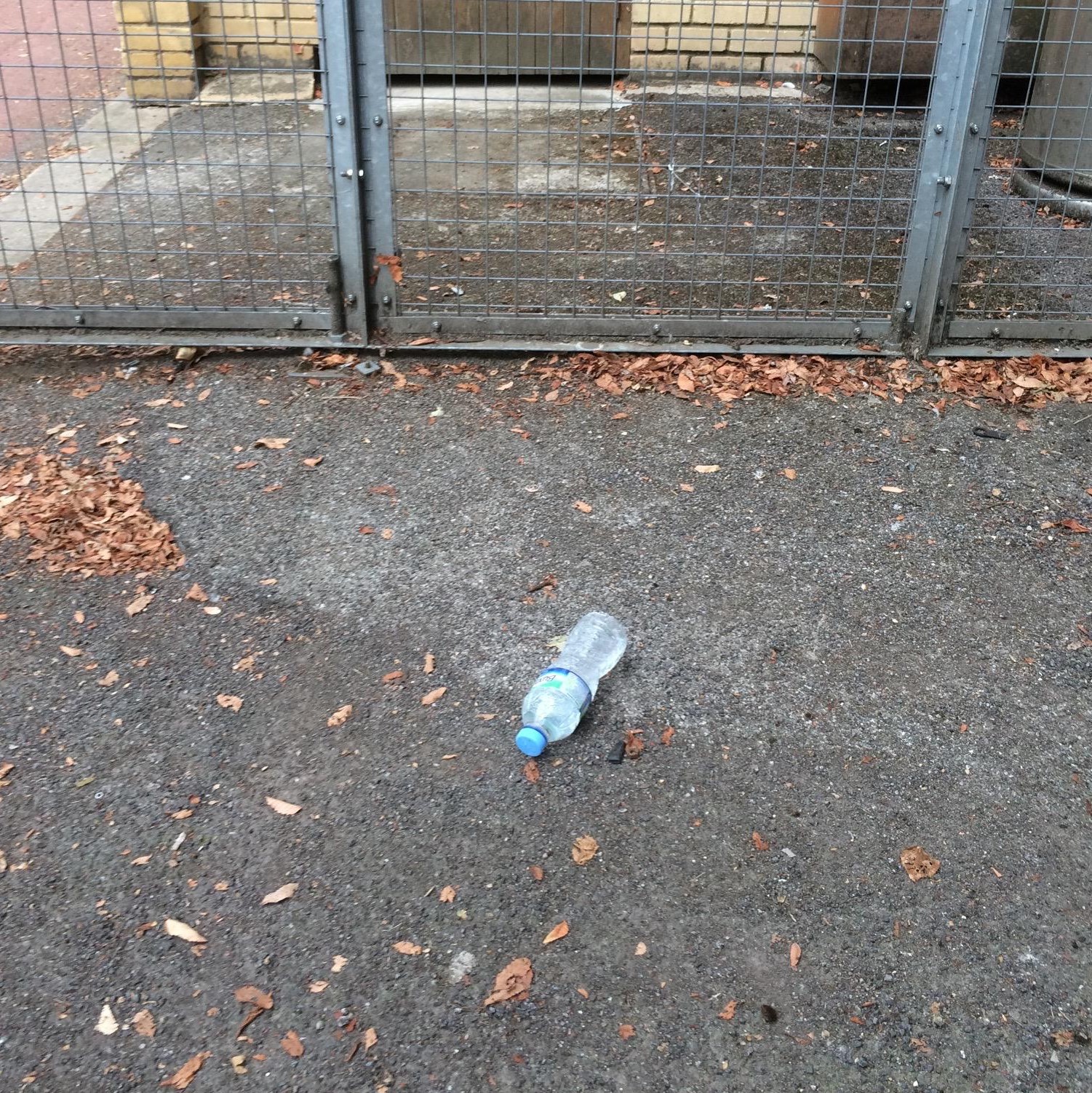}
    \includegraphics[width=0.24\linewidth]{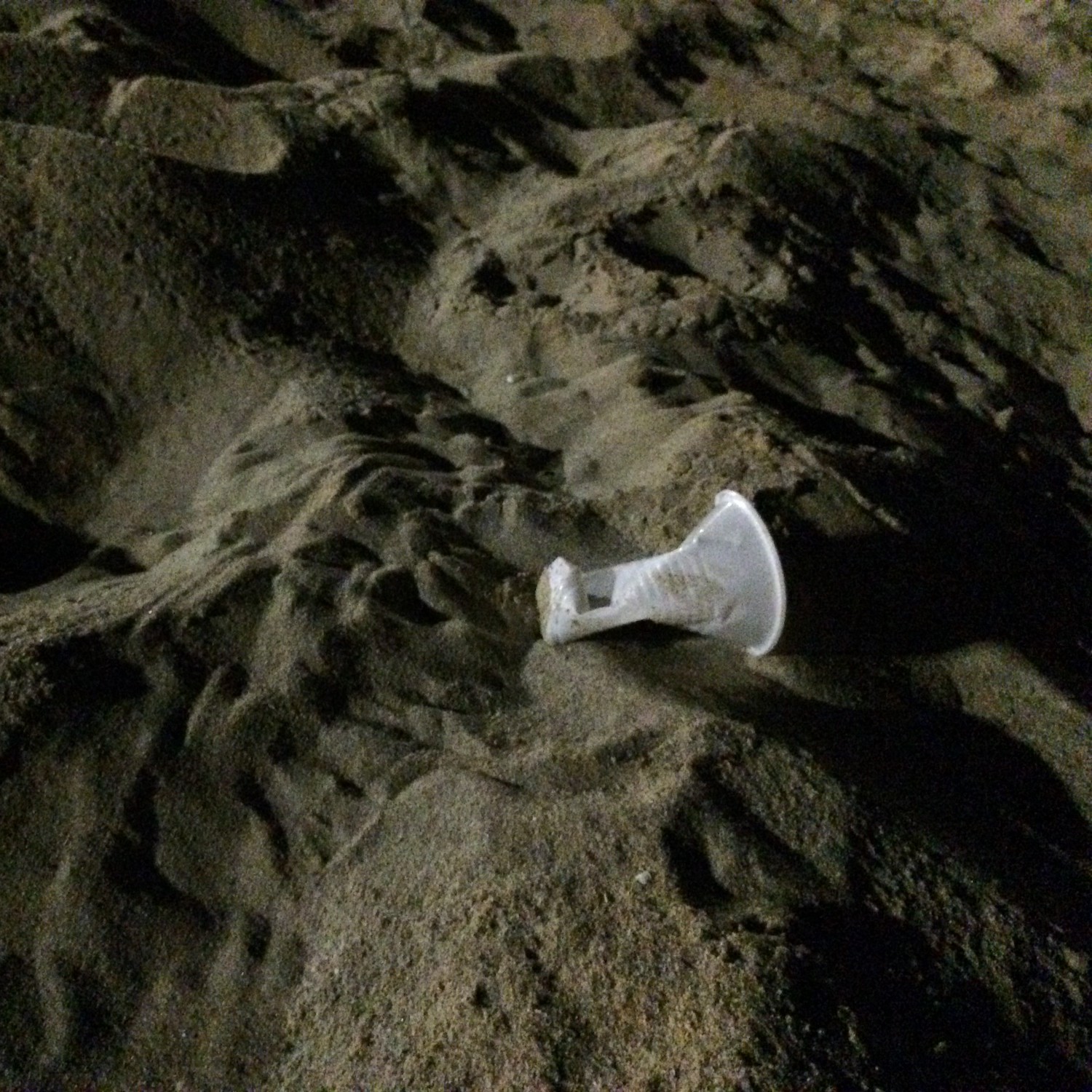}
    \caption{Example images from TACO dataset. Possible scene types are shown from left to right: vegetation, water, pavement, and sand.}
    \label{fig:taco_examples}
\end{figure}

Figure \ref{fig:taco_examples} shows examples from the TACO data set, demonstrating four types of scenes and naturally occurring occlusions. 

\subsection{Confusion Matrices}\label{app:confusion_matrices}
We include the confusion matrices corresponding to the experiments featuring alternate training methodologies, shown in Subsection \ref{subsec:alternate_training}. Keys of confusion matrices are clustered by scene for analysis of scene-specific errors, shown in Table \ref{table:distribution_percentages}.
\\ \\
\begin{minipage}[b]{.46\textwidth}
  \scriptsize
  \centering
  \begin{tabular}{l c c c}
    \hline
    & \textcolor{green}{Cupboard} & \textcolor{red}{Counter} & \textcolor{blue}{Refrig.}   \\
    \midrule
    \textcolor{blue}{Carton Soymilk} & 0\%  & 0\%  & \textcolor{blue}{100\%}   \\
    \textcolor{blue}{Can Soymilk} & 0\%  & 12\%  & \textcolor{blue}{88\%}   \\
    \textcolor{blue}{Carton OJ} & 0\%  & 12\%  & \textcolor{blue}{88\% }\\
    \textcolor{blue}{Juicebox} & 0\% & 22\%  & \textcolor{blue}{78\%} \\
    \textcolor{blue}{Diet Coke} & 0\%  & 33\%  & \textcolor{blue}{68\%} \\
    \textcolor{red}{Rice Tuscan} & 38\%  & \textcolor{red}{62\%}  & 0\% \\
    \textcolor{red}{HC Potroast} & 16\%  & \textcolor{red}{60\%}  & 24\% \\
    \textcolor{red}{Rice Pilaf} & 41\%  & \textcolor{red}{59\%}  & 0\%  \\
    \textcolor{red}{Can Chowder} & 48\% & \textcolor{red}{52\%}  & 0\% \\
    \textcolor{green}{Can Tomatosoup} & \textcolor{green}{61\%}  & 39\%  & 0\% \\
    \bottomrule
    \\
  \end{tabular}
  \captionof{table}{Clustered scene likelhoods Occluded Groceries dataset.}
  \label{table:distribution_percentages}
\end{minipage}\qquad
\begin{minipage}[b]{.46\textwidth}
  \scriptsize
  \centering
  \begin{tabular}{l c c c}
    \hline
    & \textcolor{green}{Pavement} & \textcolor{red}{Vegetation} & \textcolor{blue}{Sand...}  \\
    \midrule
        \textcolor{blue}{Styrofoam} & 29.1\% & 32.6\% & \textcolor{blue}{38.3\%}  \\
        \textcolor{red}{Can} & 30.2\% & \textcolor{red}{45.5\%} & 24.3\% \\
        \textcolor{red}{Bottle} & 30.2\% & \textcolor{red}{41.1\%} & 28.7\%  \\
        \textcolor{red}{Plastic bag..} & 31.2\% & \textcolor{red}{37.3\%} & 31.4\%\\
        \textcolor{red}{Bottle cap} & 31.8\% & \textcolor{red}{34.9\%} & 33.3\%  \\
        \textcolor{green}{Cigarette} & \textcolor{green}{50.6\%} & 21.4\% & 28.1\% \\
        \textcolor{green}{Paper} & \textcolor{green}{40.1\%} & 31.3\% & 28.6\%  \\
        \textcolor{green}{Carton} & \textcolor{green}{37.1\%} & 34.5\% & 28.3\% \\
        \textcolor{green}{Cup} & \textcolor{green}{34.8\%} & 33.9\% & 31.2\% \\
        \textcolor{green}{Straw} & \textcolor{green}{36.3\%} & 28.0\% & 35.7\% \\
        \bottomrule
        \\
      \end{tabular}
  \captionof{table}{Clustered scene likelihoods on TACO litter detection dataset.}
  \label{table:distribution_percentages_TACO}
\end{minipage}

\begin{figure}[h!]
    \includegraphics[width=0.43\linewidth]{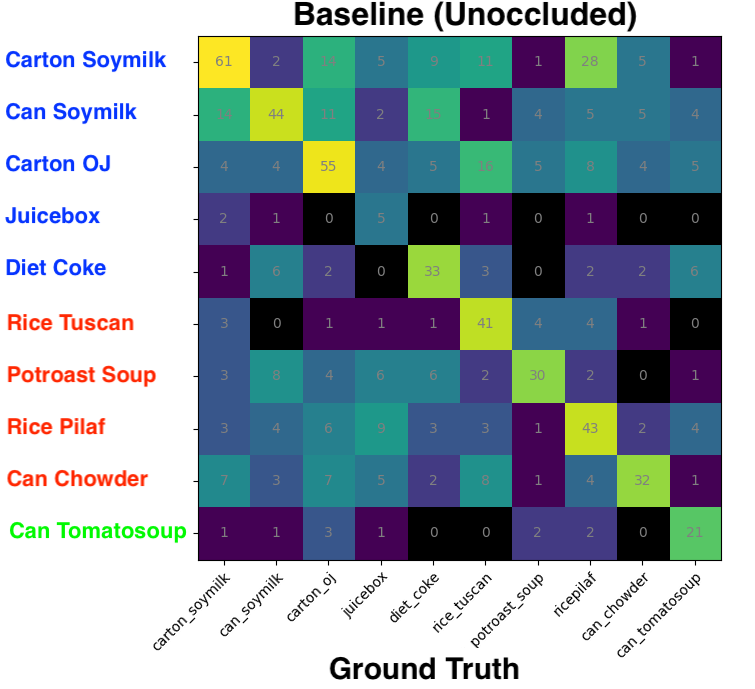}
    \hfill
    \includegraphics[width=0.43\linewidth]{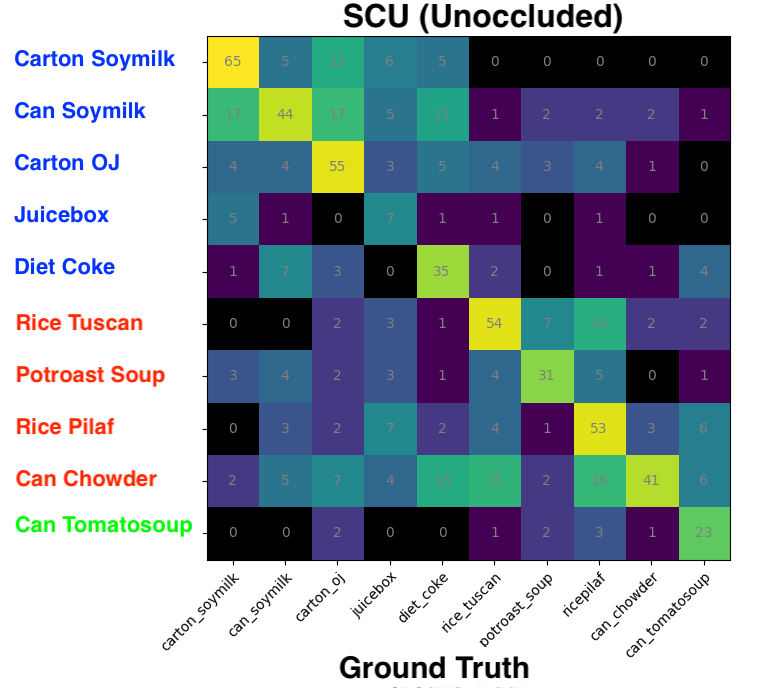}
    \\
    \caption{Confusion matrices produced from models trained on unoccluded images corresponding to Table \ref{table:unoccluded_weights}.}
\end{figure}\label{fig:confusion_matrix_unoccluded}
 
\begin{figure}[h!]
    \includegraphics[width=0.43\linewidth]{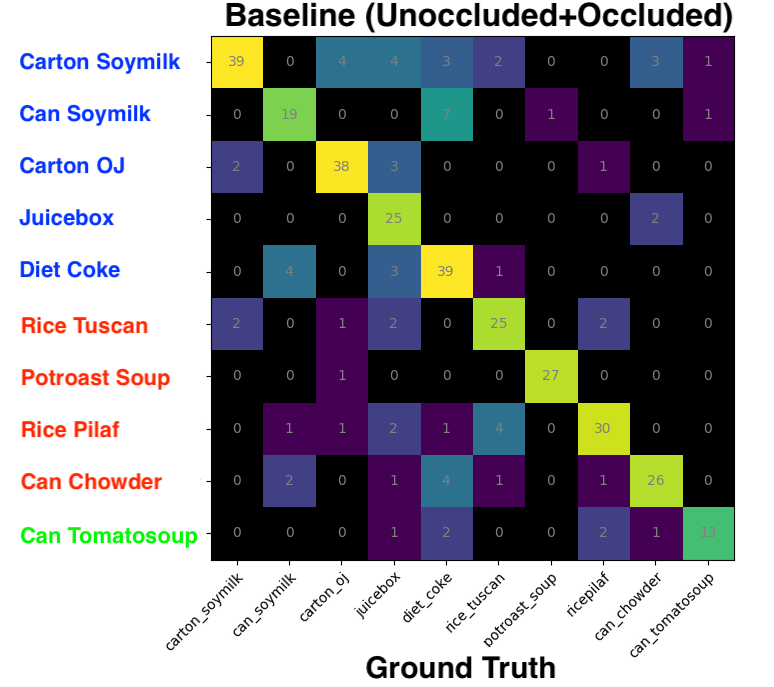}
    \hfill
    \includegraphics[width=0.43\linewidth]{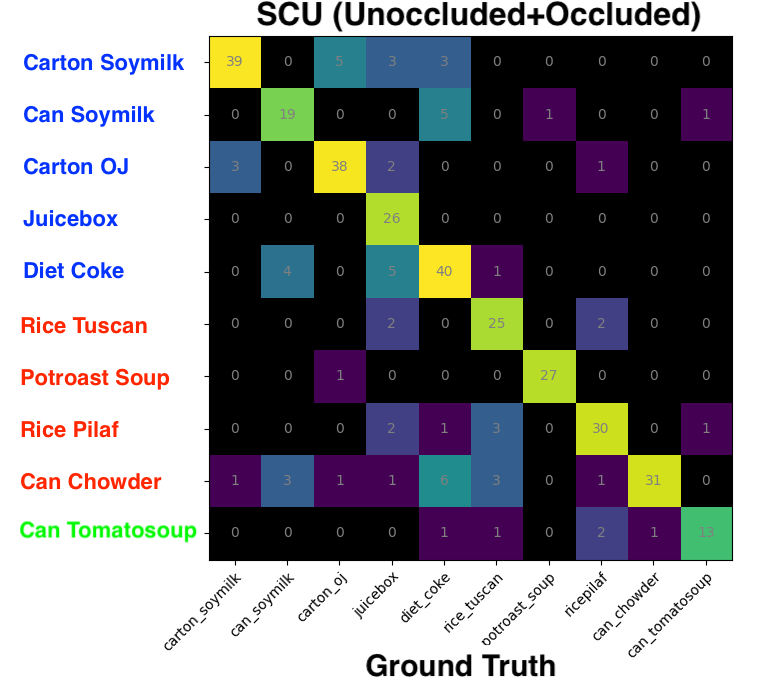}
    \\
    \caption{Confusion matrices produced from models trained on unoccluded+occluded images corresponding to Table \ref{table:unoccluded_occluded_weights}.}
\end{figure}\label{fig:confusion_matrix_unoccluded_occluded}

\subsection{Precision-Recall Curves}\label{app:pr_curves}
\begin{figure}[h!]
    \includegraphics[width=0.48\linewidth]{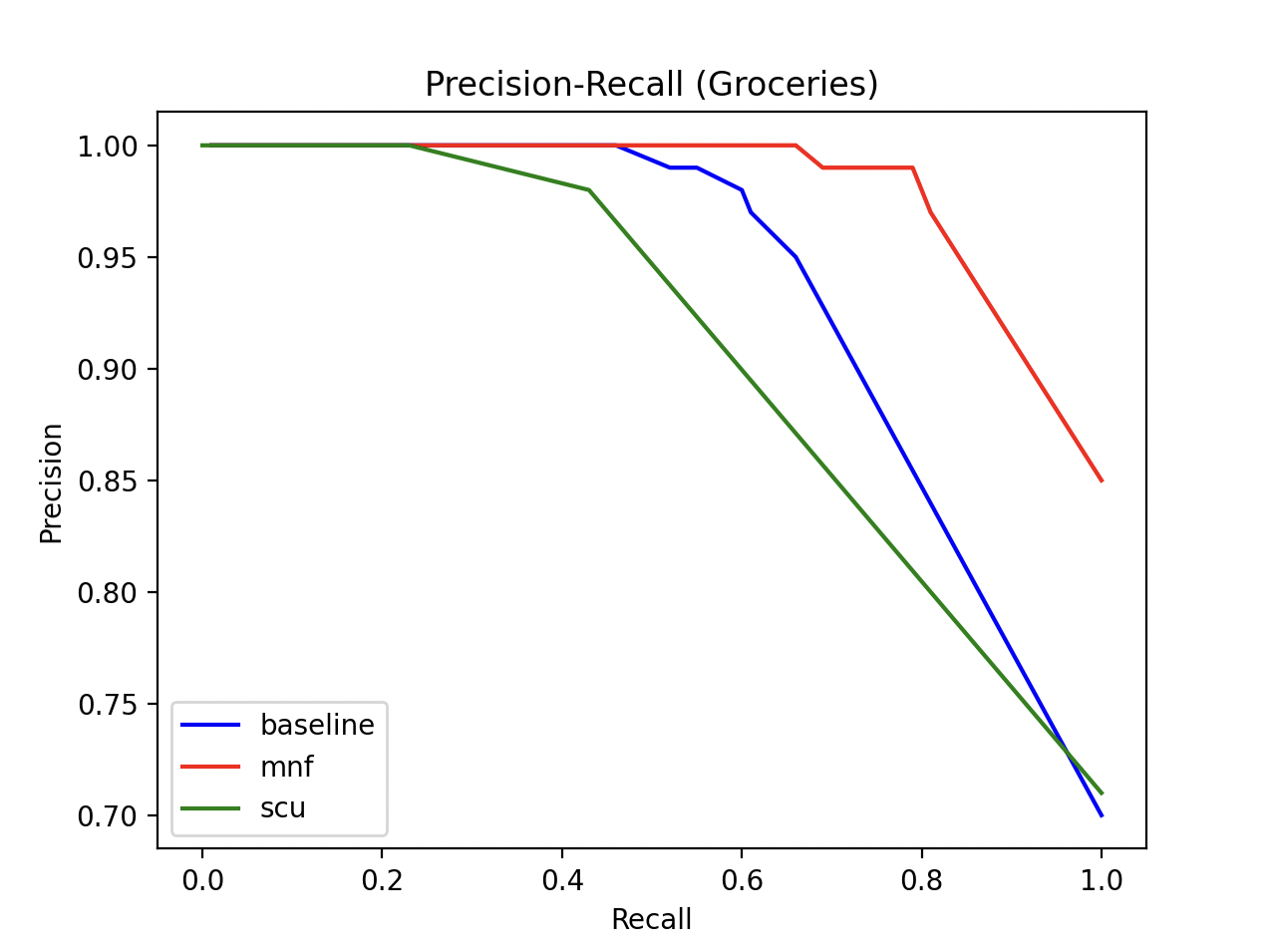}
    \hfill
    \includegraphics[width=0.48\linewidth]{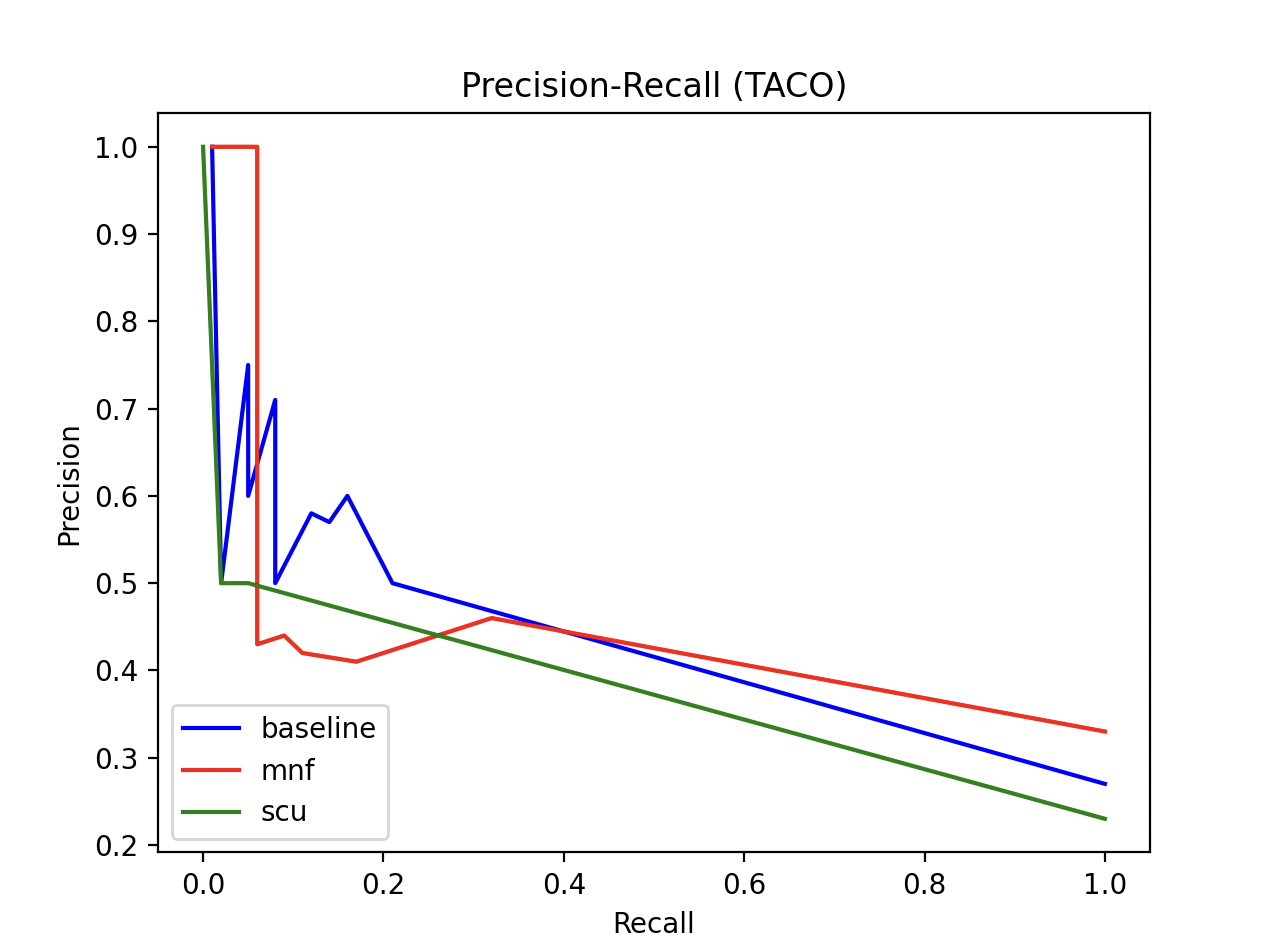}
    \\
    \caption{Precision-Recall Curves for both context algorithms against the baseline YOLO model corresponding to the results Table \ref{table:occluded_weights} and \ref{table:taco_results}.}
\end{figure}\label{fig:pr_curves}
\end{document}